\documentclass[lettersize,journal]{IEEEtran}
\usepackage{amsmath,amsfonts}
\usepackage{algorithmic}
\usepackage{algorithm}
\usepackage{array}
\usepackage[caption=false,font=normalsize,labelfont=sf,textfont=sf]{subfig}
\usepackage{textcomp}
\usepackage{stfloats}
\usepackage{url}
\usepackage{verbatim}
\usepackage{graphicx}
\usepackage{longtable}
\usepackage{cite}
\usepackage{hyperref}
\usepackage{multirow}
\usepackage{booktabs}
\usepackage{booktabs}
\usepackage{caption}
\begin{document}

\title{Zero-Shot Automatic Annotation and Instance Segmentation using LLM-Generated Datasets: Eliminating Field Imaging and Manual Annotation for Deep Learning Model Development}

\author{%
Ranjan Sapkota$^{*}$,
Achyut Paudel,
Manoj Karkee$^{*}$\thanks{$^{*}$Center for Precision and Automated Agricultural Systems, Department of Biological Systems Engineering, Washington State University, USA. Corresponding authors: \texttt{ranjan.sapkota@wsu.edu}, \texttt{manoj.karkee@wsu.edu}}
}

\maketitle

\begin{abstract}
Currently, deep learning-based instance segmentation for various applications (e.g., Agriculture) is predominantly performed using a labor-intensive process involving extensive field data collection using sophisticated sensors, followed by careful manual annotation of images, presenting significant logistical and financial challenges to researchers and organizations. The process also slows down the model development and training process. In this study, we presented a novel method for deep learning-based instance segmentation of apples in commercial orchards that eliminates the need for labor-intensive field data collection and manual annotation. Utilizing a Large Language Model (LLM), we synthetically generated orchard images and automatically annotated them using the Segment Anything Model (SAM) integrated with a YOLO11 base model. This method significantly reduces reliance on physical sensors and manual data processing, presenting a major advancement in "Agricultural AI". The synthetic, auto-annotated dataset was used to train the YOLO11 model for Apple instance segmentation, which was then validated on real orchard images. The results showed that the automatically generated annotations achieved a Dice Coefficient of 0.9513 and an IoU of 0.9303, validating the accuracy and overlap of the mask annotations. All YOLO11 configurations, trained solely on these synthetic datasets with automated annotations, accurately recognized and delineated apples, highlighting the method's efficacy. Specifically, the YOLO11m-seg configuration achieved a mask precision of 0.902 and a mask mAP@50 of 0.833 on test images collected from a commercial orchard. Additionally, the YOLO11l-seg configuration outperformed other models in validation on 40 LLM-generated images, achieving the highest mask precision and mAP@50 metrics. In terms of computational efficiency,  the YOLO11n-seg model achieved the fastest inference speed (compared to all tested configurations) at 3.8 ms among all tested configurations. These results confirm the potential of using synthetic datasets and zero-shot learning to train robust instance segmentation models, enhancing AI deployments in agriculture with improved scalability and efficiency. This method offers a viable alternative to conventional instance segmentation techniques, reducing the need for sensors, and extensive field image collection and labor-intensive manual annotation efforts while maintaining high accuracy in commercial orchard environments.  
\end{abstract}

\begin{IEEEkeywords}
Zero-shot Automatic Annotation, Automatic Segmentation , Segment Anything Model , YOLO11 , SAM , YOLO-SAM , Machine Learning , Deep Learning , Automation
\end{IEEEkeywords}

\section{Introduction}
Instance segmentation, a crucial computer vision technique that integrates object detection and semantic segmentation, provides a foundation for various automated or robotic solutions in a wide range of industries including Manufacturing, Transportation, Defense, Healthcare and Agriculture\cite{hafiz2020survey}. For example, advancement in medical imaging analysis \cite{pillai2021utilizing}, surgical planning with accurate organ mapping, disease diagnostics through detailed examination of skin lesions, blood cells, pathology samples, and dental X-ray analysis \cite{galic2023machine, brahmi2024automatic, hasnain2024idd} have been enabled by precise instance segmentation.  In the transportation systems, precise instance segmentation is a key for smarter traffic management through vehicle and pedestrian monitoring, which is a key for ensuring safety, and enhancing efficiency\cite{zhang2020traffic, wan2022edge, zhang2020virtual}. Other example applications in transportation include automating parking space allocations, improving pedestrian safety \cite{gonzalez2015review, jimenez2022perception} and optimizing railway\cite{guo2021automatic, wei2023rtlseg}, and airport operations \cite{wang2024valnet, chen2023bars, zhang2023automatic}.

Likewise, instance segmentation has been a key in advancing retail operations through enhanced customer engagement and operational efficiency \cite{gupta2024revolutionizing}. It enables real-time monitoring of product placements and inventory \cite{marder2015using, santra2019comprehensive}, optimizes store layouts through customer interaction analysis \cite{griva2018retail,wolniak2024digital}, and improves theft detection \cite{qin2021detecting, erlina2023yolo}, thereby significantly enhancing both the shopping experience and store management. Furthermore, instance segmentation techniques are crucial for improving quality control and operational efficiency \cite{islam2024deep, saini2024evaluation} in manufacturing. The techniques can be used to facilitate verification of component assembly\cite{lv2024ar, fan2024instance, yu2024novel}, ensure packaging integrity \cite{liu2024dsn,islam2024deep}, detect surface defects\cite{tabernik2020segmentation, meng2022visual, tang2023review}, guide robotic operations for better navigation and task execution \cite{khan2021vision, jiang2022review}, and refine precision machining practices  \cite{ren2022state, wang2020automatic}, all of which contribute substantially to  product quality while minimizing waste and operational costs. In addition, instance segmentation is instrumental in security surveillance, as it facilitates the accurate identification and tracking of individuals or objects over space and time \cite{tseng2021person, kang2022application}. Instance Segmentation aids in the precise identification and tracking of individuals in crowded environments \cite{lyssenko2021instance, minaee2021image} as well, which helps detecting unauthorized activities through anomaly detection \cite{sultani2018real}, and automating vehicle recognition for traffic surveillance \cite{de2022bounding, guo2021automatic}.

In agriculture, precise instance segmentation techniques are essential to advance automated and robotic operations for various field operations to achieve high productivity and produce quality in a sustainable manner. Instance segmentation can provide detailed and precise representation and localization of plant structures, which facilitates the detailed analysis of plant growth and health, impacting everything from yield estimation to disease management to robotic crop management \cite{hafiz2020survey}. Zhang et al. 2020 (\cite{zhang2020applications, zhang2020efficient}) highlight its utility in robust crop monitoring, which is essential for developing targeted interventions to control pests and manage water and nutrients. Similarly, Champ et al. (2020) \cite{champ2020instance} emphasized the importance of precise instance segmentation to innovate robotic solutions for various agricultural practices such as the thinning of immature fruits, and pollinating crop flowers showcasing its applicability in real-world agricultural operations. Likewise, past studies such as \cite{champ2020instance} have shown how convolutional neural networks (CNNs) can help reliably distinguish crops and weeds, offering a viable alternative to traditional herbicide broadcasting through targeted weed management. 

In recent years, various instance segmentation techniques have been used in developing novel solutions to address agricultural challenges. \cite{sapkota2024comparing} demonstrated the capability of deep learning-based instance segmentation to manage high variability in orchard images and to improve operational efficiencies in orchard environments. \cite{liu2019cucumber} utilized instance segmentation to enhance cucumber detection in greenhouses, while \cite{dolata2021instance} and \cite{perez2020fast} developed systems for real-time root crop yield estimation and efficient strawberry picking, respectively. Additionally, \cite{xu2023instance} and \cite{de2021dealing} improved weed management and center pivot irrigation systems detection using instance segmentation in remote sensing data. Furthermore, \cite{jia2021foveamask} introduced a model called FoveaMask for robust green fruit segmentation in instance segmentation. These studies demonstrate the broad applicability and effectiveness of instance segmentation technologies in diverse agricultural settings. Concluding these advancements, \cite{charisis2024deep} reviewed deep learning approaches in instance segmentation, emphasizing their accuracy and robustness in crop stress and growth monitoring. 

Instance segmentation often faces significant challenges due to the limited availability of labeled data, which typically requires extensive and labor-intensive data preparation and manual annotation. Addressing this challenge, few-shot learning has gained prominence as a method that trains models to perform tasks with a minimal amount of labeled examples, marking a shift from conventional methods. Few-shot learning enables the model to learn from a small number of training samples, reducing the dependency on extensive datasets and mitigating the need for laborious manual annotation.\cite{fan2020fgn} introduced the Fully Guided Network (FGN), a novel architecture that combines few-shot learning with Mask R-CNN. FGN employs tailored guidance mechanisms within Mask R-CNN to optimize inter-class generalization, significantly outperforming previous state-of-the-art methods by leveraging a support set. Additionally, \cite{ganea2021incremental} developed an incremental few-shot instance segmentation model, iMTFA, which utilizes discriminative embeddings to add new classes efficiently without the need for retraining, while also minimizing memory utilization by storing embeddings rather than images. This method greatly enhances scalability and adaptability. Moreover, \cite{han2024reference} proposed Reference Twice (RefT), a unified transformer-based framework that prevents overfitting and boosts the few-shot instance segmentation process through simple cross-attention mechanisms that connect support and query features. Their approach achieved substantial performance improvement across various settings, deminstrating the effectiveness of their class-enhanced base knowledge distillation loss in bridging the gap between resource-heavy traditional methods and more efficient, scalable machine learning techniques.

Recent studies have demonstrated significant strides in enhancing model adaptability with limited data through few-shot learning which is a technique that trains models to predict accurately with minimal labeled data, a shift from traditional methods that need large datasets \cite{wang2020generalizing, perez2021true}. This technique uses only a few examples per class, making it ideal where collecting extensive data is impractical \cite{sun2021research}. It includes approaches like Siamese networks for sample comparison, meta-learning for quick task adaptation, and transfer learning to fine-tune pre-trained models on small datasets. Recent advancements have notably enhanced model adaptability with limited data through these methods. \cite{keaton2023celltranspose} addressed automated cellular instance segmentation by utilizing specialized contrastive losses that effectively leverage a minimal number of annotations. This approach significantly mitigated covariate shift effects, demonstrating that few-shot learning can produce competitive results even with limited data, reducing the dependence on extensive datasets and computational resources. Similarly, \cite{wang2022dynamic} developed the Dynamic Transformer Network (DTN), which uses Dynamic Queries conditioned on reference images to directly segment target instances, eliminating the need for dense proposal generation and extensive post-processing. This end-to-end method improves optimization and generalization in few-shot instance segmentation, with a Semantic-induced Transformer Decoder that effectively minimizes background noise, leading to significant improvements when tested on the COCO-20 dataset. Furthermore, \cite{nguyen2022ifs} extended the Mask-RCNN framework with iFS-RCNN for incremental few-shot instance segmentation, introducing a probit-based classifier and an uncertainty-guided bounding box predictor. This novel approach uses Bayesian learning to handle scarce data for new classes and incorporates advanced loss functions.

Further advancing these models, various innovative methodologies have been developed in recent years for  zero-shot learning approach \cite{belissent2024transfer, ge2025alignzeg, kim2024generalized}, which performs instance segmentation without reliance on labeled data for unseen classes. \cite{zheng2021zero} proposed a comprehensive approach for Zero-Shot Instance Segmentation (ZSI), which is an emerging area in deep learning that aims to segment and identify object instances in images without having seen examples of those specific object classes during training. Their approach integrates a Zero-shot Detector, Semantic Mask Head, Background Aware RPN, and a Synchronized Background Strategy, establishing a new benchmark for ZSI on the MS-COCO dataset with promising results. Meanwhile, \cite{shin2023zero} introduced Zero-shot Unsupervised Transfer Instance Segmentation (ZUTIS), a framework that operates without instance-level annotations and achieves good model performance on standard datasets. Additionally, \cite{bucher2019zero} tackled the challenge of scaling semantic segmentation to a large number of object classes with their ZS3Net, which leverages deep visual models and semantic word embeddings to classify pixels for unseen categories during testing. This method sets baselines on standard segmentation datasets like Pascal-VOC and Pascal-Context. \cite{ren2023visual} provided an overview of visual semantic segmentation based on few/zero-shot learning, describing the rapid innivation and advancement in this area. They discussed the integration of few/zero-shot learning with 2D and 3D visual semantic segmentation, and highlighted the potential and challenges in expanding these techniques to practical applications.

\begin{figure*}[ht]
\centering
\includegraphics[width=0.80\linewidth]{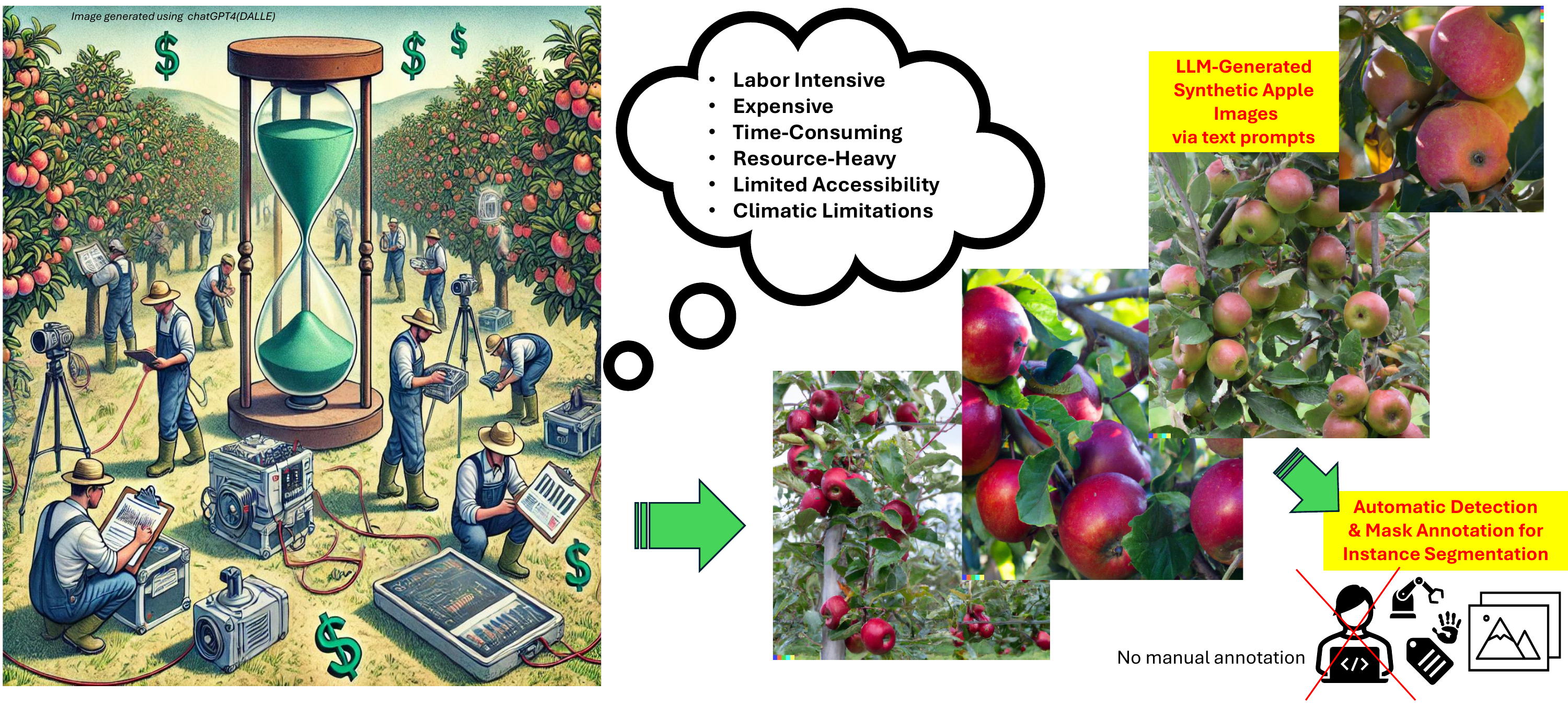}
\caption{Showing a contrast in data collection methods for instance segmentation in agriculture. On the left, human workers use sophisticated sensors to collect images from orchards and engage in manual labeling, illustrating the traditional, labor-intensive process. On the right, the use of LLMs simplifies this process by generating and automatically annotating realistic images of orchards, showcasing an efficient approach.}
\label{fig:ProblemStatement}
\end{figure*}
Despite the advancements in few-shot and zero-shot learning techniques, deep learning-based instance segmentation in numerous application areas still predominantly relies on a labor-intensive process of data collection from actual field environments and manual labeling  (Figure \ref{fig:ProblemStatement}). The current approach incurs significant logistical and financial challenges in field data collection, as well as in acquiring and/or developing sophisticated sensors, and in labor-intensive manual annotation, which is substantial barrier to scalability and practical implementation. Instance segmentation traditionally incurs high costs due to the need for advanced sensors and their maintenance, skilled operation, and extensive travel for field data collection.  Manual annotation of the collected data is time-consuming, error-prone, and costly due to the expertise required. The factors significantly increase logistical and financial burdens, complicating the deployment and scalability of deep learning applications not only in agriculture but across various fields. 

Recently, few-shot learning has emerged as a promising approach, reducing the dependency on extensive labeled datasets by training models to perform with limited data exposure. However, the transition to zero-shot learning, where models recognize and segment objects without prior specific training examples, could further reduce the reliance on these extensive and costly data collection and annotation processes. This approach could be especially impactful where data scarcity and rapid adaptability are crucial (e.g., Agriculture). These technologies have the potential to eliminate the need for any dataset during training, allowing for robust model deployment in variable environments (e.g., orchards) without prior exposure to target data.

Additionally, recent advancements in language models have revolutionized synthetic data generation, demonstrating significant potential across various application areas, including agriculture, healthcare, and digital art \cite{sapkota2024synthetic, liao2024text, chen2024twigma, cho2023dall, vayadande2023ai}. The ability of LLMs to generate photorealistic images from textual descriptions has been harnessed to create training datasets where acquiring real-world data is impractical due to high costs or logistical challenges \cite{sapkota2024synthetic}. In our recent study, we utilized OpenAI's DALL-E to generate and automatically annotate images of orchards, streamlining the training of machine vision models like YOLOv10 and YOLO11 for apple detection in agriculture \cite{sapkota2024synthetic}. This approach not only reduced the need for field data collection but also achieved high accuracy when validated against real orchard images, highlighting the efficacy of LLM-generated datasets in practical settings. \cite{liao2024text} explored the use of LLMs in converting abstract concepts into tangible visual representations through a Text-to-Image generation framework for Abstract Concepts (TIAC). This method clarified abstract concepts into detailed visual forms, enabling the creation of images that effectively communicate complex ideas without physical examples. \cite{chen2024twigma} introduced TWIGMA, a dataset of generative AI images, to analyze the distinct characteristics and public engagement with AI-generated content. This study provided insights into the evolving preferences and acceptance of synthetic images among users, indicating a growing reliance on generative models for creating diverse visual content. \cite{cho2023dall} assessed the capabilities and biases of text-to-image models, revealing limitations in visual reasoning and social bias concerns that suggest areas for future improvement. Despite generating high-quality images, these models often struggled with accurately depicting object relationships and maintaining fairness across different demographics. \cite{vayadande2023ai} developed a web application utilizing DALL-E to generate images based on user descriptions, demonstrating the model’s practical utility in interactive digital environments. This application enabled users to create custom visuals effortlessly, showcasing the model's adaptability and user-friendly nature. Despite these advancements, a significant gap remains in the field of deep learning-based instance segmentation, where the process still heavily relies on labor-intensive data collection and manual annotation. The use of LLMs to generate fully annotated datasets presents a promising avenue to mitigate these challenges. By automating both the generation and annotation of images, LLMs could  reduce the cost and time required for developing and training deep learning models in many fields (such as agriculture) where variability and data accessibility are major hurdles. 

In this study, we address the typical challenges of deep learning-based instance segmentation techiques by eliminating the need for physical sensor-based data collection and manual annotation. We introduce a workflow that leverages large language models (LLMs) to generate synthetic imagery, accurately replicating real-world conditions in commercial apple orchards (Figure \ref{fig:ProblemStatement} ). This approach reduces logistical and financial constraints, improving the scalability and applicability of instance segmentation across various domains. By adopting a fully automated process that includes zero-shot detection and mask segmentation, our method offers a robust solution that can be replicated for different objects of interest in numerous applications, marking a significant advancement in automated image processing. Specific objectives of the study are:
\begin{itemize}
    \item Utilize LLM-generated images to develop a custom deep-learning model without the need for physical data collection. 
    \item Apply zero-shot detection using a YOLO11 model coupled with the Segment Anything Model (SAMv2) to perform automatic mask segmentation and annotation of apples.
    \item Train a custom YOLO11 instance segmentation model using an automatically annotated, synthetic dataset (obj2) and assess its precision and accuracy in detecting apples.
    \item Validate the trained model against in-field images collected from a commercial orchard using a machine vision sensor to demonstrate the model’s effectiveness and practical applicability.
\end{itemize}

\section{Methods}
\begin{figure*}[ht]
\centering
\includegraphics[width=0.87\linewidth]{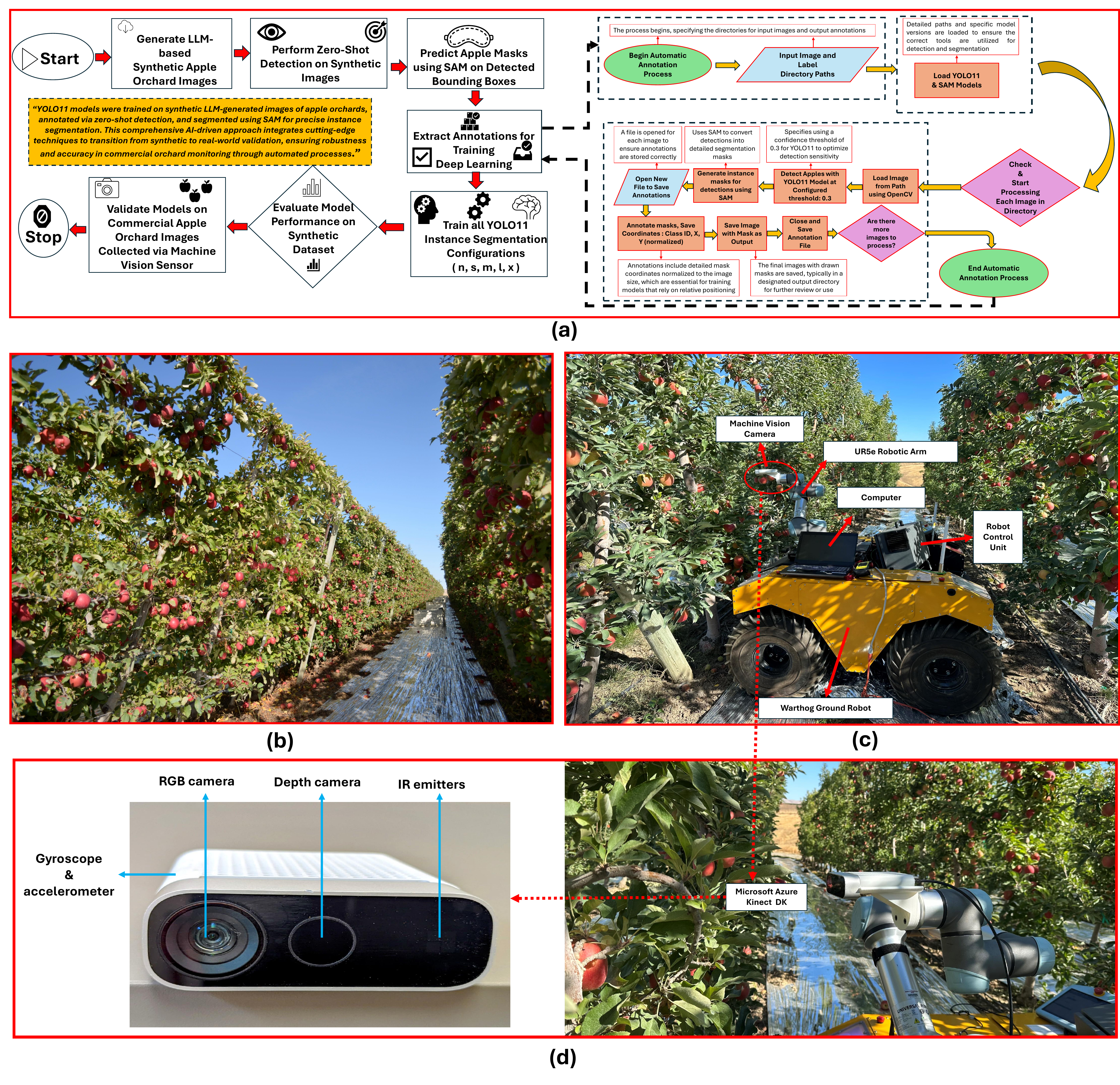}
\caption{a) Process diagram illustrating the development of a deep learning model for generation and automated annotation of synthetic apple tree images without the use of physical sensors, field data collection, or manual annotations; b) A sample image of the commercial "Scifresh" apple orchard in Prosser, Washington State, USA, where the developed model was validated, demonstrating the practical application of the synthetic image generation and automated annotation methods; c) A sensing system (including a ground robot) used as the platform for image collection in the orchard; d) Microsoft Azure Kinect DK machine vision camera used for model validation using real-world images, demonstrating the model's applicability in sensor-based, real-world systems.}
\label{fig:OverallMethod}
\end{figure*}

The proposed method of developing and evaluating an instance segmentation method in commercial apple orchards utilizing synthetically generated and automatically labeled images is depicted  in Figure \ref{fig:OverallMethod}a. The method began with utilizing the realistic apple orchard images generated by the LLM (DALL.E), a multi-modal LLM, recognized for producing high-fidelity visual content. Subsequently, these images underwent a zero-shot learning with a YOLO11 object detection model trained only on COCO benchmark dataset that identified apples by creating bounding boxes around them. An image processing technique was then developed to estimate apple boundaries using the Segment Anything model \cite{kirillov2023segment} for semantic segmentation within the bounding boxes created by zero-shot YOLO11 model. Apple boundary annotations generated in this process were automatically saved in .txt format, forming a foundational dataset for further model training. This dataset was then used to train the YOLO11 instance segmentation model, specifically tailored to improve the accuracy of apple segmentation. To validate the model's performance in practical orchard environments, it was tested against real-world images from commercial orchards, captured using a machine vision system. Furthermore, for a comprehensive comparison of automatically generated masks against manually annotated ground truths, the LLM-generated images were subjected to both manual and automatic annotation using the proposed method. An additional dataset collected by a machine vision sensor from a commercial apple orchard was similarly annotated using both manual and automated processes and compared to assess the performance of the proposed method. 

In summary, the methodology employed in this research included five sequential stages: 1) Generation of Synthetic Images Using Large Language Models (LLMs), 2) Zero-Shot Detection using YOLO11, 3) Automated Mask Annotation Generation using SAMv2, 4) Training YOLO11 Instance Segmentation Models, and 5) Performance Evaluation and Validation in a Commercial Apple Orchard.  

\subsection{Study Site and Data Acquisition}
The study site for the validation of the deep learning model developed for instance segmentation of apples was a commercial 'Scifresh' apple orchard located in Prosser, Washington State, USA. The data acquisition occurred in October 2024, during the peak harvest season, as depicted in Figure \ref{fig:OverallMethod}b. This orchard was characterized by tree rows spacing of 10 feet and individual tree spacing of 3 feet (Figure 2b). Although no imaging was utilized during the development phase of the model, validation was conducted on-site as shown in Figure 2b. Images were captured using a machine vision camera (Microsoft Azure Kinect DK \ref{fig:OverallMethod}d), which was mounted on a robotic platform, detailed in Figure \ref{fig:OverallMethod}c. The robotic platform comprised a Universal Robotic arm Ur5e affixed to a ground robot Warthog, provided by Clearpath Robotics of Ontario, Canada.
The study site for the validation of the deep learning model developed for instance segmentation of apples was a commercial 'Scifresh' apple orchard located in Prosser, Washington State, USA. The data acquisition occurred in October 2024, during the peak harvest season, as depicted in Figure \ref{fig:OverallMethod}b. This orchard was characterized by tree rows spacing of 10 feet(3 meters approx) and individual tree spacing of 3 feet (Figure 2b). Although no imaging was utilized during the development phase of the model, validation was conducted on-site as shown in Figure 2b. Images were captured using a machine vision camera (Microsoft Azure Kinect DK \ref{fig:OverallMethod}d), which was mounted on a robotic platform, detailed in Figure \ref{fig:OverallMethod}c comprised of a Robotic arm (Ur5e, Universal Robotics) installed on a ground platform (Warthog, Clearpath Robotics of Ontario, Canada).

\subsection{Generation of Synthetic Images Using Large Language Model}
In this study, the synthetic image generation process utilized the DALLE model to create realistic orchard scenarios with text prompts without physical data collection, as detailed in our previous study \cite{sapkota2024synthetic}. A total of 524 images from our previous study \cite{sapkota2024synthetic} (as shown in right side of Figure \ref{fig:ProblemStatement}) depicting various states and formations of apples were generated based on succinct, precise textual prompts, and each image measured 1024 by 1024 pixels. These prompts efficiently guided the DALL.E model to produce visually accurate orchard scenes, and the images generated were rigorously reviewed to ensure realism and relevance for subsequent analysis. The LLM-generated dataset can be found in github link : https://github.com/ranzosap/Synthetic-Meets-Authentic. 

\subsection{Zero-Shot Detection using YOLO11}
The model was applied to detect apples without prior exposure to a specific apple dataset, employing a method known as zero-shot detection \cite{bansal2018zero, zhu2019zero}. This machine learning approach enables a model to recognize and categorize objects that it has not explicitly been trained on. Conventionally, machine learning models require extensively annotated datasets to learn object features for effective detection. Zero-shot learning, however, leverages the model’s ability to generalize from known categories to novel ones without direct training examples \cite{pourpanah2022review, nilforoshan2023zero, chen2023zero}.

\begin{figure*}[ht]
\centering
\includegraphics[width=0.88\linewidth]{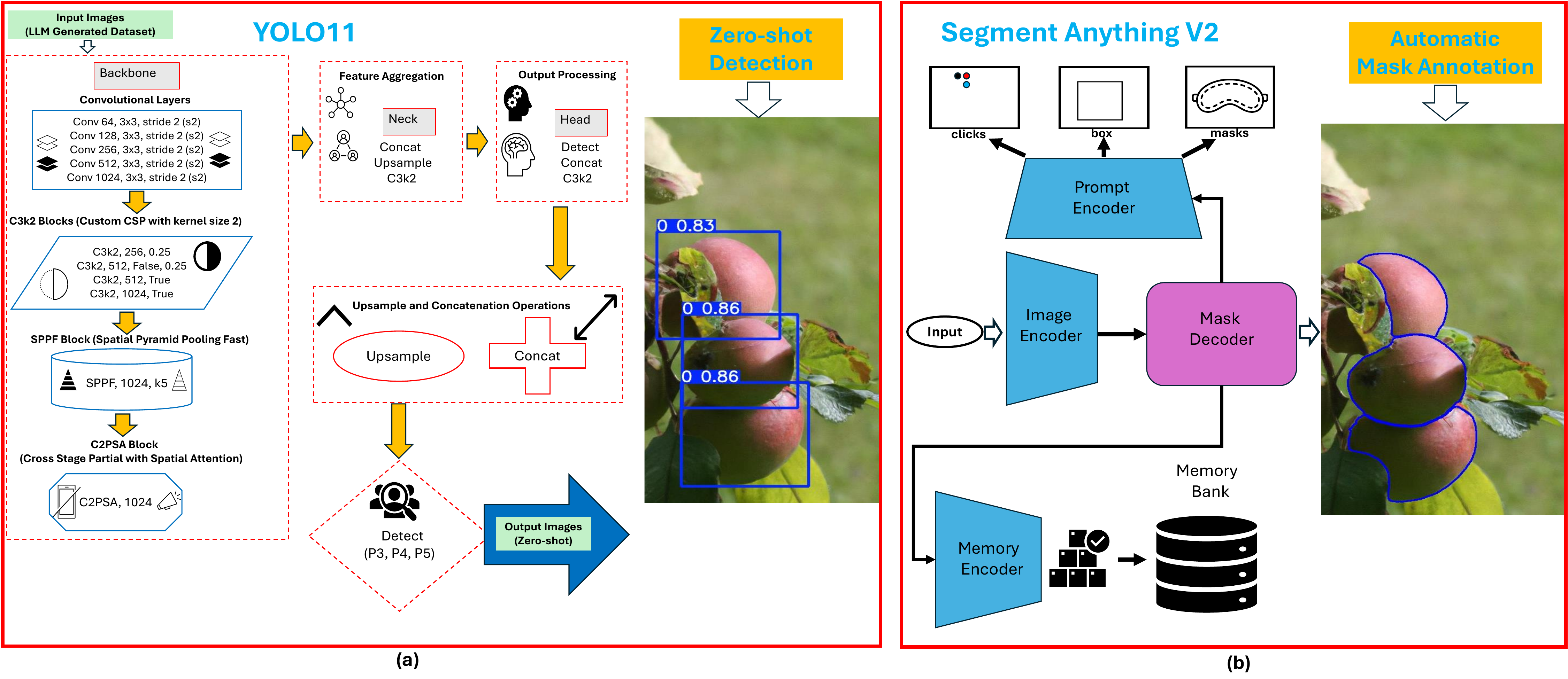}
\caption{Overview of the automatic annotation process using YOLO11 and SAM models: a) Zero-shot detection using YOLO11 applied to synthetic orchard images, illustrating the model's capability to identify apple instances;  and b) Subsequent automatic mask annotation using SAM}
\label{fig:YOLOSAMANNOTATION}
\end{figure*}

The YOLO11 base model (zero-shot) identified the apples as region of interest in bounding box within LLM-generated images, as shown in Figure\ref{fig:YOLOSAMANNOTATION}a. This process involved no manual annotations, demonstrating the model's ability to leverage learned detection principles from previously trained categories and apply them to new objects like apples, highlighting its advanced feature extraction capabilities. The model was executed using an Intel Xeon\textregistered{} W-2155 CPU at 3.30 GHz, NVIDIA TITAN Xp Collector’s Edition graphics card, and 31.1 GiB of memory on an Ubuntu 16.04 LTS 64-bit system. This zero-shot detection model generated bounding boxes of apples to localize them in images (Figure\ref{fig:YOLOSAMANNOTATION}a). The bounding box for each detected object is defined as:
\begin{equation}
    B = (x, y, w, h),
\end{equation}

where $(x, y)$ represents the center of the box and $(w, h)$ denotes its dimensions. YOLO11 calculates these parameters using feature maps derived from the images, scaled by factors learned during training on diverse object sets. The bounding boxes, autonomously generated by YOLO11, encapsulated the detected apples, showcasing the model's capacity in executing zero-shot detection of apples in commercial orchard environment.

\subsection{Automatic Generation of Mask Annotation using Segment Anything Model (SAMv2)}
As a foundational model, SAM \cite{kirillov2023segment} uses masked autoencoders for robust image encoding, coupled with a versatile prompt encoder that handles various input types such as points, boxes, masks, and text \cite{xiong2024efficientsam, zhang2024segment}. This model is distinctive for its ability to generate precise segmentation masks based on minimal input, demonstrating a powerful zero-shot capability that requires no prior training on specific objects \cite{osco2023segment, maquiling2024zero}. SAM's architecture is built around several core components: a high-capacity image encoder that uses a vision transformer to process images into dense embeddings, and a prompt encoder that interprets sparse and dense prompts \cite{kirillov2023segment}. The image encoder is particularly notable for its use of a ViT-H/16 vision transformer, which manages to efficiently downscale images to a more manageable size while retaining critical visual information. The prompt encoder handles a range of input types, converting them into formats that can be effectively used to modify image embeddings. The mask decoder is where SAM's innovative segmentation capabilities are realized \cite{osco2023segment, dong2024efficient}. This component integrates image and prompt embeddings to produce final segmentation masks. It utilizes a Transformer-based decoder architecture that includes self-attention and cross-attention mechanisms, allowing it to refine the segmentation output in response to the input prompts. To train such a sophisticated model, the creation of a suitable dataset was a considerable challenge. The development team employed a novel model-in-the-loop annotation system that evolved from assisted-manual annotations to fully automated mask generation. This approach not only streamlined the dataset creation process but also ensured that SAM could be trained on a vast array of images with corresponding high-quality masks, facilitating its zero-shot transfer capabilities. In this study SAMv2 was used to generate precise mask annotations for instance segmentation of apples using the apple bounding boxes identified by zero shot YOLO11  (Figure \ref{fig:YOLOSAMANNOTATION}a) (Figure \ref{fig:YOLOSAMANNOTATION}b).

Following the zero-shot detection by the YOLO11 model, the methodology for automatic annotation of apples was executed using the SAMv2. Initially, the YOLO11 model, equipped with pre-trained weights from a prior training session, was used to detect objects within the LLM-generated orchard images. Subsequently, the SAMv2 model, loaded with its specific pre-trained weights, was applied to generate segmentation masks (Figure \ref{fig:OverallMethod}a). The integration of these two models achieved the automatic creation of segmentation masks, delineating each apple within the images without the need for manual annotation.

This step begins by retrieving LLM-generated images of apple orchards from a designated directory followed by individually processing them with YOLO11 for object detection (apples in this case), applying a confidence threshold (0.3) to ensure the precision of detected objects. Detected apples were enclosed within bounding boxes, whose coordinates were provided as inputs to the SAM model for generating the detailed segmentation masks for each identified apple. The detected objects were enclosed within bounding boxes, which served as input for the SAMV2 model to generate semantic masks. These masks were normalized relative to image dimensions, saved in label files following the YOLO11 format, and used for consistent annotation and training in the dataset as depicted in Figure \ref{fig:YOLOSAMANNOTATION}b. This step was performed on a workstation with an Intel Xeon® W-2155 CPU @ 3.30 GHz x20 processor, NVIDIA TITAN Xp Collector's Edition/PCIe/SSE2 graphics card, 31.1 GiB memory, and Ubuntu 16.04 LTS 64-bit operating system.

\subsection{Evaluation of Automatic Annotation Method}
As discussed before, the mask annotation method was developed using YOLO and SAM models. Each image was first subjected to object detection via the YOLO model, which identified potential apple regions (bounding boxes) with a minimum confidence threshold of 0.3. These bounding boxes served as inputs for the SAM2 model, which then estimated segmentation masks within the defined regions. The taken for each step of the preprocessing, inference, and postprocessing steps were recorded for assessing the computational efficiency of the model.

\subsubsection{Metrics and Statistical Evaluation}
The quantitative evaluation involved computing specific metrics across the dataset to gauge the model's performance. These metrics included were the average number of detections per image and the mean confidence of detections, defined as:

\begin{align}
    N_d &= \frac{1}{n} \sum_{i=1}^n \text{count}(detections_i) \\
    C_{avg} &= \frac{1}{n} \sum_{i=1}^n \text{mean}(confidence_{detections_i})
\end{align}

where $n$ is the total number of images processed, $detections_i$ denotes the detections in the $i^{th}$ image, and $confidence_{detections_i}$ is the confidence level associated with each detection. These metrics provide insight into the model’s detection and segmentation capabilities within LLM-generated orchard environments. A high average confidence indicates robust detection capabilities, while the number of detections reflects the model’s precision and effectiveness in segmenting apple instances. 

Additionally, the performance of the automatic annotation technique was conducted on a set of test images which were both manually and automatically annotated for comparing those masks. 

Initially, a random selection of 40 images from a set of 501 LLM-generated images was annotated manually. This annotation process involved delineating the mask for each apple using a manual annotation tool available in Roboflow (Roboflow, IOWA, USA). Concurrently, the same images were subjected to the proposed automatic annotation process and compared with the manual annotation. Based on the manual and automatic mask annotations, Average Precision, Average Recall, Average F1-Score, Average Dice Coefficient, and Average IoU were calculated as follows. 

\begin{itemize}
    \item \textbf{Average Precision}, \textbf{Average Recall}, and \textbf{Average F1-Score} were derived from the counts of True Positives (TP), False Positives (FP), and False Negatives (FN) across all images, calculated by:
    \begin{align}
        \text{Average Precision} &= \frac{1}{N} \sum_{i=1}^N \frac{\text{TP}_i}{\text{TP}_i + \text{FP}_i} \\
        \text{Average Recall} &= \frac{1}{N} \sum_{i=1}^N \frac{\text{TP}_i}{\text{TP}_i + \text{FN}_i} \\
        \text{Average F1-Score} &= \frac{1}{N} \sum_{i=1}^N 2 \times \frac{\left(\frac{\text{TP}_i}{\text{TP}_i + \text{FP}_i}\right) \times \left(\frac{\text{TP}_i}{\text{TP}_i + \text{FN}_i}\right)}{\left(\frac{\text{TP}_i}{\text{TP}_i + \text{FP}_i}\right) + \left(\frac{\text{TP}_i}{\text{TP}_i + \text{FN}_i}\right)}
    \end{align}
    where $N$ is the total number of images in the dataset.

    \item \textbf{Average Dice Coefficient} and \textbf{Average IoU} were calculated by averaging the respective metrics for each pair of predicted and ground truth masks across all images:
    \begin{align}
        \text{Average Dice Coefficient} &= \frac{1}{N} \sum_{i=1}^N \frac{2 \times | \text{TP}_i |}{2 \times | \text{TP}_i | + | \text{FP}_i | + | \text{FN}_i |} \\
        \text{Average IoU} &= \frac{1}{N} \sum_{i=1}^N \frac{|\text{A}_i \cap \text{B}_i|}{|\text{A}_i \cup \text{B}_i|}
    \end{align}
    where $A_i$ and $B_i$ represent the automatic and manual mask areas for the $i^{th}$ image, respectively.
\end{itemize}

These metrics together provided a comprehensive assessment of the model’s ability to perform instance segmentation without manual intervention, emphasizing the reduction in time and labor typically required for dataset collection and preparation in deep learning applications.

\subsection{Training the YOLO11 Instance Segmentation Model}

\begin{figure*}[ht]
\centering
\includegraphics[width=0.89\linewidth]{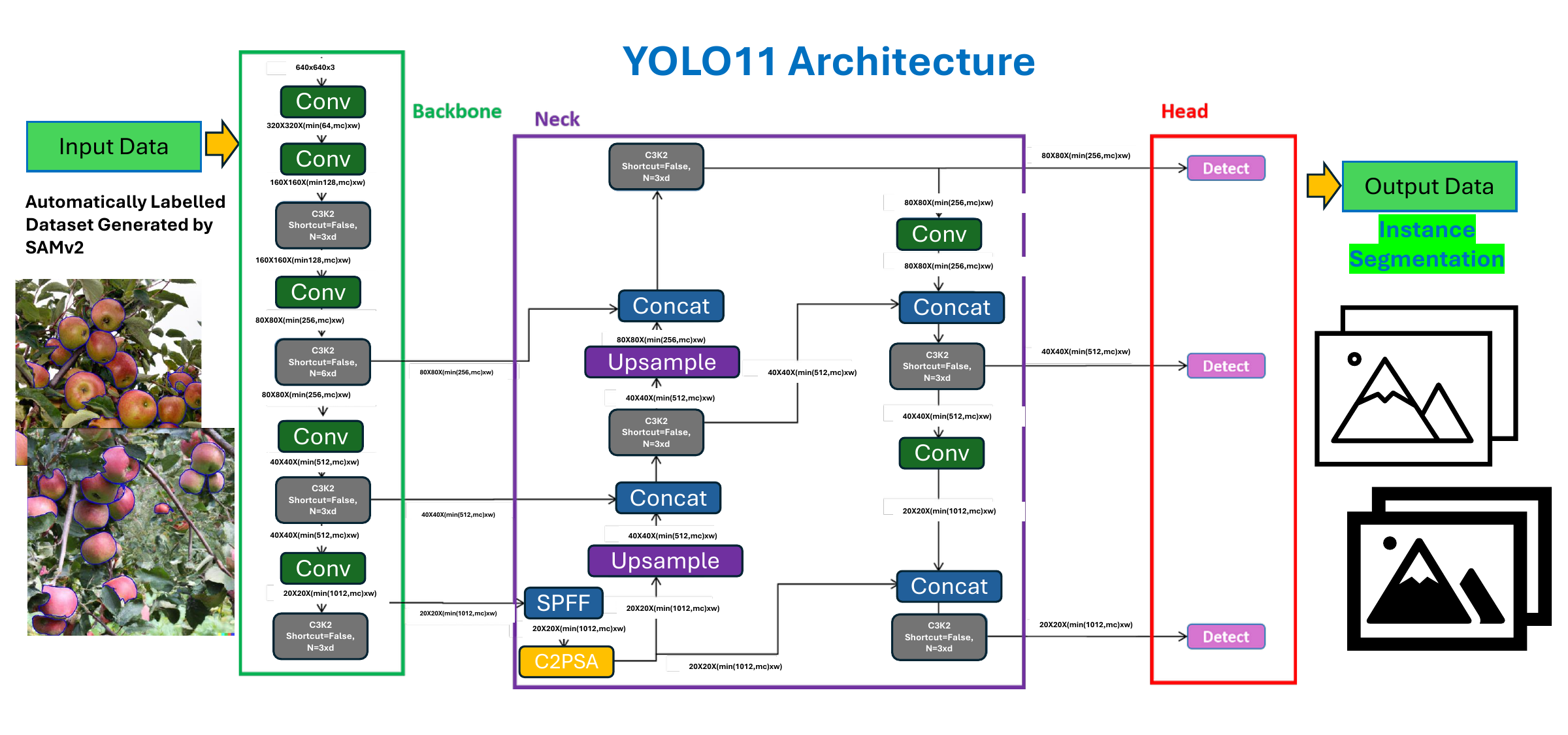}
\caption{YOLO11 model architecture used for detection and segmentation of apples in commercial orchards using LLM-generated and automatically annotated dataset for training}
\label{fig:yolo11Architecture}
\end{figure*}

Once mask annotations were created automatically as discussed above, various configurations of YOLO11 instance segmentation models (Models used in recent studies \cite{sapkota2024yolo11, sapkota2024comparing, sapkota2024comprehensive, sapkota2024synthetic}) were fine-tuned for apple instance segmentation using the automatically generated annotations as shown in Figure \ref{fig:yolo11Architecture}. The five model configurations used were: YOLO11n, YOLO11s, YOLO11m, YOLO11l, and YOLO11x, which were  trained under identical hyperparameter settings to ensure consistency and comparability across different model scales. The models were trained using a uniform batch size of 8 and an image resolution of 640x640 pixels. This standardization was crucial for maintaining consistency in how each model learned from the synthetic image dataset. The training employed an automatic optimizer (adam) and was conducted over 300 epochs while patience was set to 100 to avoid overfitting..

Key hyperparameters were carefully selected to optimize the training outcomes. The initial learning rate was set at 0.01, with momentum at 0.937 and weight decay at 0.0005, which are parameters that influence the speed and stability of the learning process. Additionally, a warmup phase of 3 epochs was included to stabilize the learning parameters before entering the main training phase. Specific loss settings were also applied, including a box loss of 7.5 and a class loss of 0.5, which helped fine-tune the precision of the models in identifying and classifying objects within images. Image augmentation techniques such as flipping, translation, and a mosaic were utilized to enhance the robustness of the models against variations in real-world scenarios. These techniques simulate different imaging perspectives and fruit occlusions, providing the models with a broader range of data to learn from. As discussed before, the models were trained exclusively on synthetic images generated by a Large Language Model (LLM) and annotated automatically using a zero-shot YOLO11 and SAMv2 models. This approach leverages a pre-trained base model for initial detections without further training on our dataset, emphasizing the experiment's innovative use of synthesized data and zero-shot learning techniques. 

The model was trained using  an Intel Xeon(R) W-2155 CPU with a base clock speed of 3.30 GHz and 20 cores, alongside NVIDIA Corporation GP102 [TITAN Xp] graphics cards. The system also featured 7.0 TB of storage and operated on Ubuntu 20.04.6 LTS with a GNOME version 3.36.8 graphical interface and X11 windowing system.

\subsection{Performance Evaluation of YOLO11 Instance Segmentation}
The efficacy of instance segmentation was assessed by calculating Mask Precision, Recall, and mean Average Precision (mAP) at a 50\% Intersection over Union (IoU) threshold (mAP@50). These metrics were derived based on the overlap between the predicted masks and the ground truth annotations. Mask Precision, which quantifies the proportion of correctly identified positive predictions, is defined as:

\begin{equation}
    \text{Mask Precision} = \frac{\text{TP}}{\text{TP} + \text{FP}}
\end{equation}

Mask Recall measures the ability to detect all relevant instances:

\begin{equation}
    \text{Mask Recall} = \frac{\text{TP}}{\text{TP} + \text{FN}}
\end{equation}

mAP@50 represents the mean AP at the 50\% IoU threshold, providing a balanced view of precision and recall across different decision thresholds.

Additional performance metrics assessed included image processing speeds for preprocessing, inference, and post-processing stages, as well as training time. The architectural complexity of each model was also assessed in terms pf the number of convolutional layers, the total number of parameters, and the computational load in gigaflops (Giga Floating Point Operations per Second or GFLOPs). The architectural complexity and computational efficiency of the YOLO11 was assessed by examining three critical aspects: the number of parameters, GFLOPs , and the count of convolutional layers utilized in each configuration. These factors are essential indicators of a model's potential performance and operational demands.

Parameters, representing the total count of trainable elements within the model, were evaluated to understand the models' complexity and memory requirements:

The model parameters are defined as:
\begin{equation}
    \text{Parameters}_{\text{Model}} = \text{Trainable weights + biases}
\end{equation}

GFLOPs, representing the computational load during the inference phase, are calculated to provide insights into the model's efficiency and speed:
\begin{equation}
    \text{GFLOPs} = \frac{\text{Total floating-point ops.}}{10^9} \text{ per image}
\end{equation}

The architectural depth, indicated by the number of convolutional layers which influences feature extraction capabilities, is given by:
\begin{equation}
    \text{Layers}_{\text{Conv}} = \text{Total conv. layers}
\end{equation}

\section{Results}
\begin{figure*}[ht]
\centering
\includegraphics[width=0.85\linewidth]{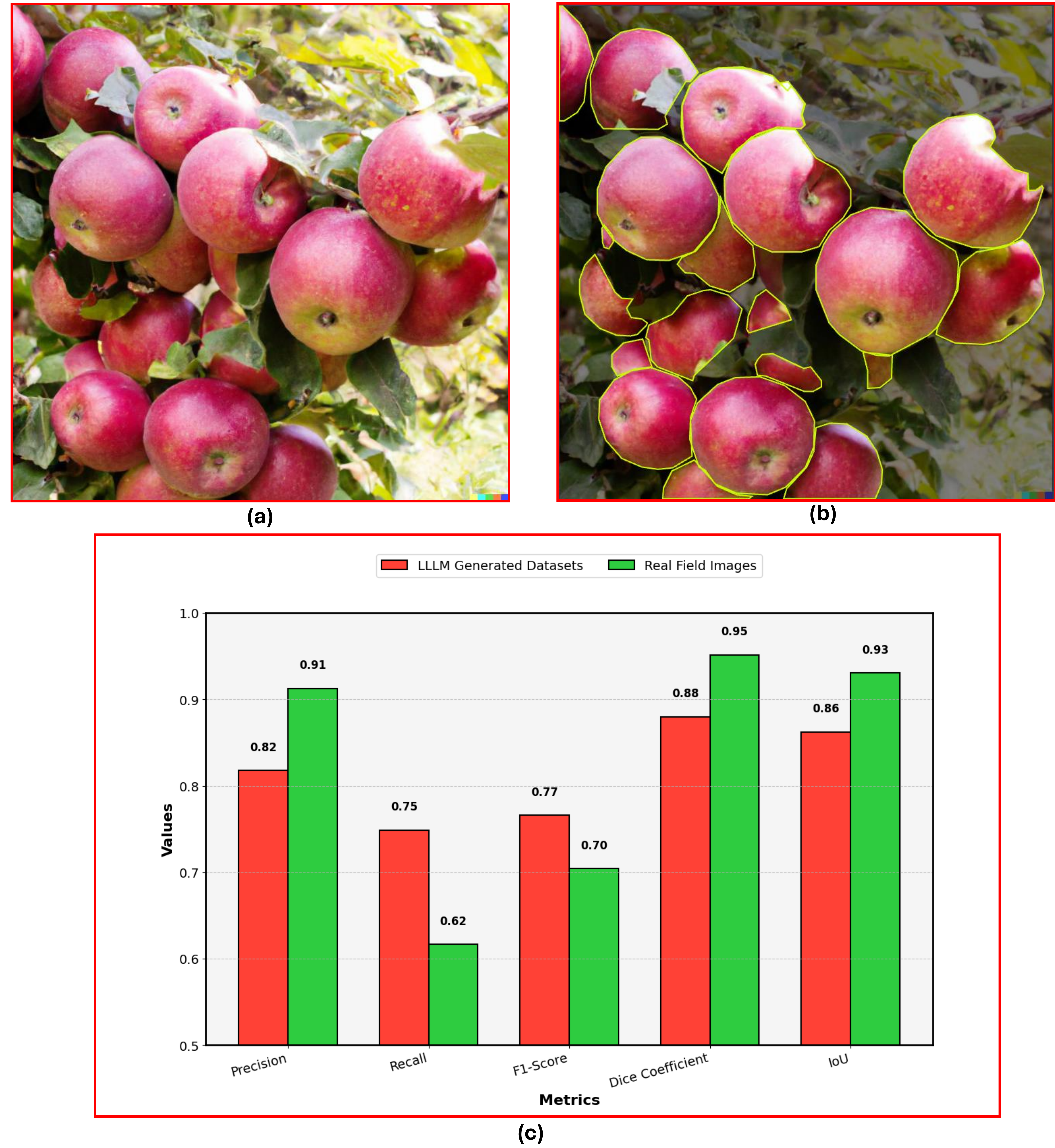}
\caption{Demonstrating effective mask annotations for deep learning-based instance segmentation model training : (a) Showing LLM-generated image of an apple orchard. (b) Showing automatic labeling post-YOLO11 base model zero-shot detection using SAM : (c)Showing the comparative performance metrics between LLLM Generated Datasets and Real Field Images for instance segmentation. Metrics such as Precision, Recall, F1-Score, Dice Coefficient, and IoU.}
\label{fig:results1}
\end{figure*}

The result of the automatic mask annotation method is presented in Figure \ref{fig:results1}, where panel Figure \ref{fig:results1}(a) shows an example apple orchard image generated by the LLM. Panel \ref{fig:results1}(b) illustrates the effectiveness of automated mask annotation following the integration of zero-shot YOLO11 model and SAMv2. The LLM-generated images show a high degree of realism, underlining the potential of our method for generating training dataset for instance segmentation models in agricultural settings. The fusion of YOLO11's zero-shot detection capabilities with SAM's precise mask-generation techniques led to promising outcomes in automated image annotation. Qualitatively, our approach successfully identified and annotated both fully and partially visible apples within the complex orchard scene, a task that would otherwise demand substantial manual labor. 
\par 
As mentioned in the methods section, the quantitative performance of the automatic mask annotation method was evaluated using manual annotations of apples in 40 synthetic and 42 real orchard images. Figure \ref{fig:results1}(c) highlighted the effectiveness of our approach for training deep learning models for instance segmentation. With the synthetic (LLM Generated) Dataset, the proposed method achieved high Dice Coefficients and IoU values of 0.88 and 0.86, respectively, demonstrating strong overlap and accuracy of mask annotations. For orchard images (Real Field Images) collected with a Microsoft Azure Kinect camera, the method reached an even higher precision of 0.91, although the recall was lower at 0.61. These metrics demonstrated the potential of our automatic annotation approach, across both synthetic (LLM-generated Images) and complex real-world environments.

Qualitatively, the automated mask annotation process detected and masked most of the apples as shown in Figure \ref{fig:results1} (b). On average, the model successfully detected and masked 8 objects (apples) per image with a high average confidence of 0.80. The computational time (particularly the inference time) per image increased with the increasing  number of target objects in those images. On average, the model took 4.5 ms, 1,986.4 ms (1.9 seconds), and 1.1 ms per image respectively for pre-processing, inference, and post-processing respectively. 

\subsection{Performance of YOLO11 Instance Segmentation Model against Automatic Annotation}
\begin{figure}[ht]
\centering
\includegraphics[width=0.95\linewidth]{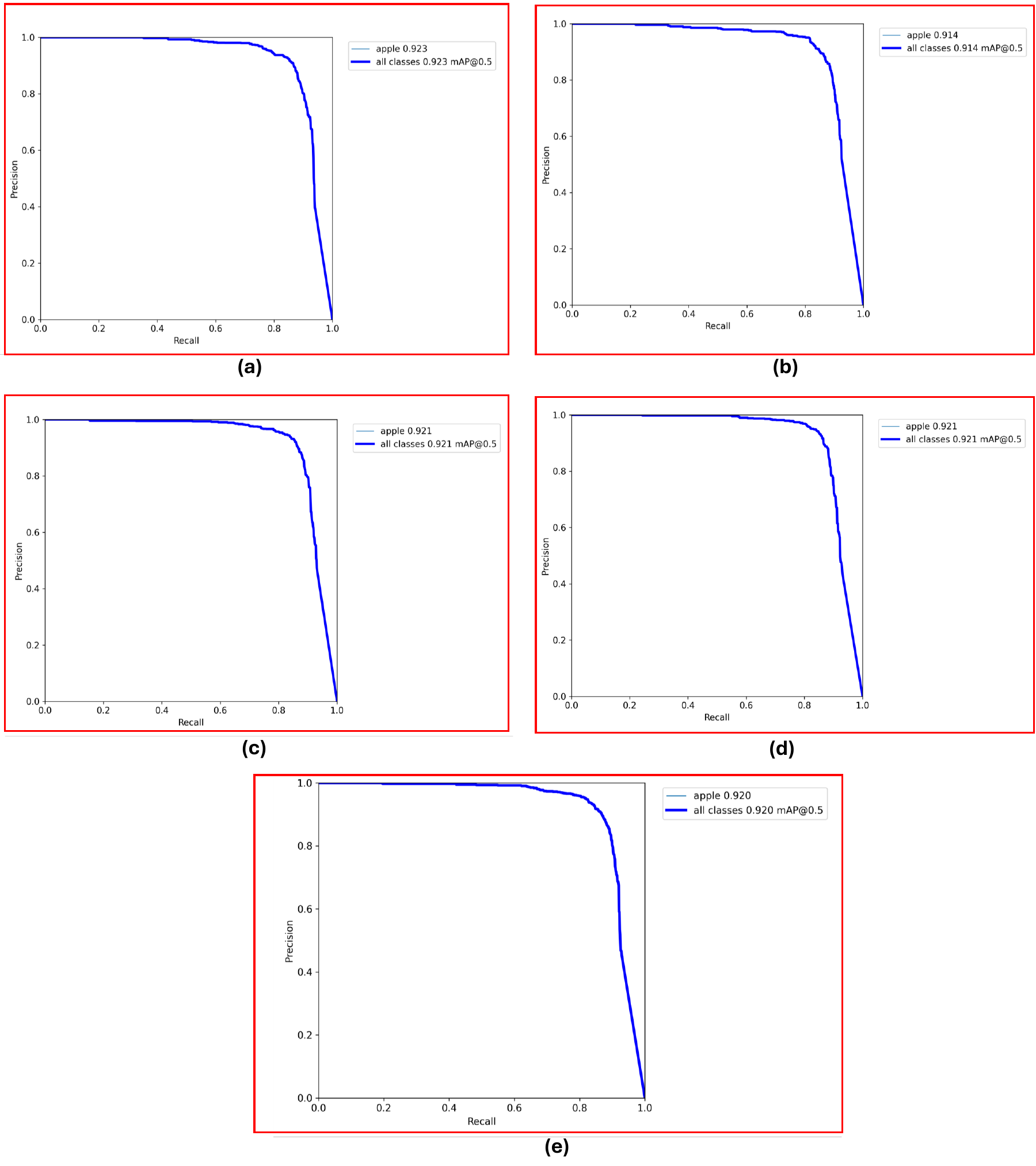}
\caption{Precision-Recall Curves of YOLO11n-seg,s-seg,m-seg,l-seg, and x-seg models on instance segmentation of apples in LLM generated and automatic annotated datasets}
\label{fig:PRcurve}
\end{figure}
Figure \ref{fig:PRcurve} shows the precision-recall curves of the five configurations of the YOLO11 instance segmentation models over the LLM-generated synthetic dataset. Each sub-figure from (a) to (e) corresponds to different model configurations (YOLO11n-seg, YOLO11s-seg, YOLO11m-seg, YOLO11l-seg, and YOLO11x-seg), showcasing mean Average Precision (mAP@0.5) values close to 0.92 across all classes. Table \ref{tab:metrics_llm} shows the detailed evaluation of box and mask metrics for the detection and instance segmentation of apples in the LLM-generated dataset. Among the trained models, the YOLO11l-seg model achieved the highest performance (See table \ref{tab:metrics_llm}) on the synthetic test dataset (40 LLM-generated images), with a mask precision of 0.93 and overall mask mAP@50 of 0.92. 

These results illustrate the efficacy of each YOLO11 configuration in handling instance segmentation on synthetic datasets of apple orchards. The YOLO11l-seg configuration stands out with the highest performance, but the other models also show commendable capabilities, each achieving a mean Average Precision (mAP@0.5) close to 0.92. This indicates a robust ability to accurately detect and segment apples across various model complexities. The nearly uniform high performance across these configurations underscores their practical applicability in agricultural settings, offering effective solutions even with limited computational resources. These results underscore the potential for deploying these models in real agricultural operations to facilitate tasks like fruit counting and disease identification without the need for extensive manual intervention. By leveraging synthetic datasets of apple orchards, agricultural producers can reduce both costs and complexity, making advanced AI technologies more accessible and feasible for widespread use in challenging and resource-sensitive environments.

\begin{table*}[h]
\centering
\caption{Evaluation of instance segmentation metrics for various configurations of YOLO11 model on LLM-generated dataset.}
\renewcommand{\arraystretch}{1.2} 
\fontsize{8}{10}\selectfont
\begin{tabular}{@{}lcccccccc@{}}
\toprule
\textbf{Model} & \multicolumn{4}{c}{\textbf{Box Metrics}} & \multicolumn{4}{c}{\textbf{Mask Metrics}} \\
& \textbf{P} & \textbf{R} & \textbf{mAP50} & \textbf{mAP50:95} & \textbf{P} & \textbf{R} & \textbf{mAP50} & \textbf{mAP50:95} \\
\midrule
YOLO11n-seg & 0.905 & 0.872 & 0.935 & 0.884 & 0.916 & 0.855 & 0.923 & 0.846 \\
YOLO11s-seg & 0.900 & 0.868 & 0.926 & 0.887 & 0.893 & 0.862 & 0.914 & 0.853 \\
YOLO11m-seg & 0.928 & 0.858 & 0.932 & 0.890 & 0.925 & 0.855 & 0.921 & 0.855 \\
YOLO11l-seg & 0.932 & 0.859 & 0.935 & 0.890 & 0.931 & 0.856 & 0.921 & 0.852 \\
YOLO11x-seg & 0.908 & 0.868 & 0.932 & 0.891 & 0.909 & 0.862 & 0.920 & 0.851 \\
\bottomrule
\end{tabular}
\label{tab:metrics_llm}
\end{table*}

\begin{figure*}[ht]
\centering
\includegraphics[width=0.88\linewidth]{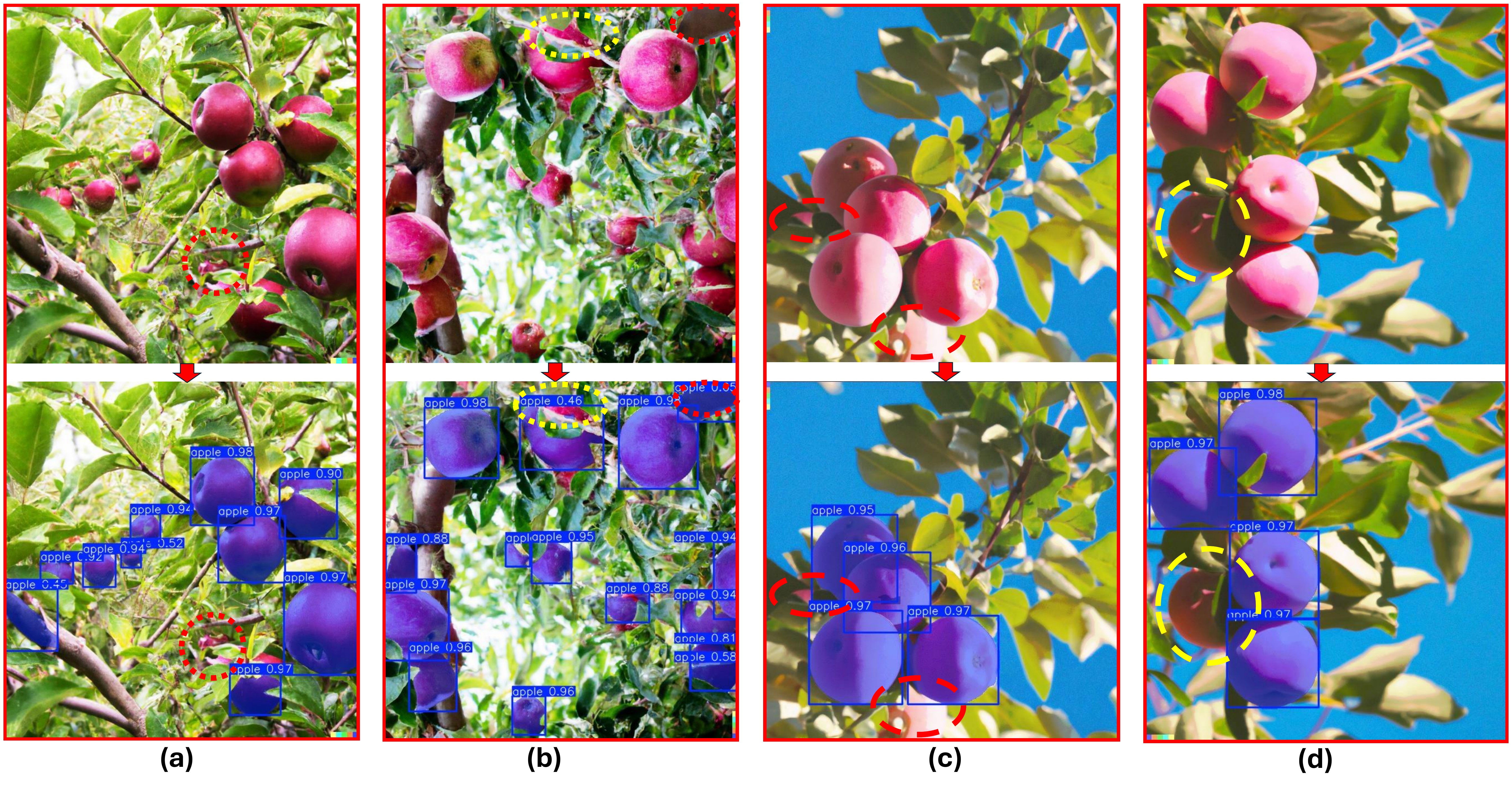}
\caption{Example output of YOLO11n-seg instance segmentation model on DALL.E-generated images with manual annotation; (a) A missed detection of an apple-like region (red dotted circle), though other apples are accurately segmented; (b) A false segmentation of a branch as an apple (red circle) and partial segmentation of an occluded apple (yellow dotted circle), reflecting challenges with similar color and texture; (c) and (d) Undetected apples in complex, occluded scenarios (red and yellow dotted circles), emphasizing the model's limitations and need for larger, more varied training datasets to improve detection.}
\label{fig:resultsYOLO11LLMgenerated}
\end{figure*}
Figure \ref{fig:resultsYOLO11LLMgenerated} provides an example of qualitative performance of YOLO11n-seg instance segmentation on highly realistic images generated by the DALL.E model. The top section of the figure displays LLM-generated images depicting varying orchard scenarios, while the bottom section shows the outcomes of YOLO11 instance segmentation applied to these images. The qualitative successes demonstrated in Figure \ref{fig:resultsYOLO11LLMgenerated}  highlight the YOLO11n-seg model's robust capacity to accurately identify and segment apples within complex and varied orchard scenarios. For instance, as seen in \ref{fig:resultsYOLO11LLMgenerated}a, the model proficiently detected and segmented apples positioned in the background, confirming its effectiveness in handling distant objects. Moreover, Figure \ref{fig:resultsYOLO11LLMgenerated}b showcases the model's capability to consistently detect and segment nearly all visible apples, even amidst the challenging conditions of a densely populated orchard scene.  Figure \ref{fig:resultsYOLO11LLMgenerated}(a) (highlighted in red dotted line), an apple-like region in a sample LLM-generated image was not accurately identified by the YOLO11 model, though the model successfully identified and segmented other apples in the background. This example illustrates both the capabilities and limitations of applying advanced instance segmentation techniques based on automatically annotated  synthetic datasets.

\par 
Also, Figure \ref{fig:resultsYOLO11LLMgenerated} b demonstrates a challenging scenario where a shadowy branch region on the top right corner, was erroneously segmented as an apple due to its similar appearance. This misidentification underscores need for model improvement in distinguishing between actual fruit and similar-looking background elements. Conversely, an example of successful segmentation in a challenging situation is depicted by the yellow dotted region on the same figure \ref{fig:resultsYOLO11LLMgenerated} b. Here, despite the apple being partially obscured by foliage, the YOLO11 model accurately recognized and segmented the fruit. However, it is noted that the segmentation was incomplete as some portions of the apple remained to be unsegmented, another area for potential improvement in the future to achieve robust detection in challenging orchard environments. Figure \ref{fig:resultsYOLO11LLMgenerated} c highlights another challenging situation where apples were illuminated brightly and were partially occluded. These apples, visible only in segments, were not detected by the model (red dotted circles). Similarly, Figure \ref{fig:resultsYOLO11LLMgenerated}d, marked by a yellow circle, shows another instance where an apple was missed by the YOLO11 instance segmentation model. These detection challenges could potentially be mitigated by expanding the training dataset. The study, conducted with only 501 synthetic images generated via text prompts to a LLM, suggests that enlarging the dataset would be essential to potentially enhance model performance, leveraging the capability to generate huge volumes of data quickly and at low cost using LLMs. 
\begin{figure*}[ht]
\centering
\includegraphics[width=0.85\linewidth]{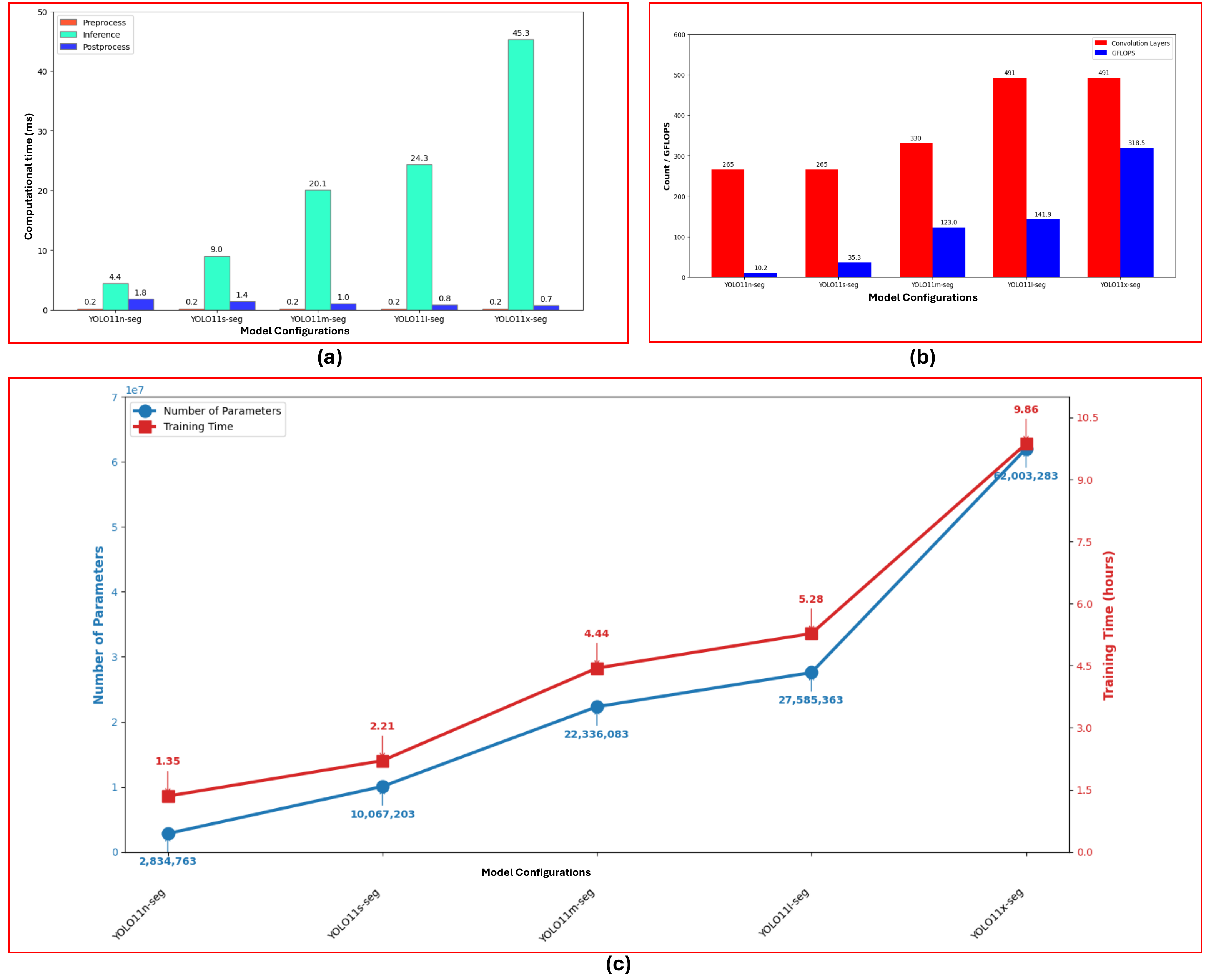}
\caption{Performance metrics of various YOLO11 configurations on LLM-generated dataset: (a) image processing speed, highlighting efficient pre-processing and varied inference; (b) Comparison of the number of convolution layers and GFLOPs, indicating the computational complexity; and (c) Comparison of training time taken by individual models (for the same 300 epochs) and the number of model parameters, reflecting their capacity and computational demand.}
\label{fig:AllREST}
\end{figure*}

As illustrated in Figure \ref{fig:AllREST}a, the various YOLO11 model configurations displayed differential computational speeds during their training and evaluation on LLM-generated datasets. Pre-processing time consistently remained at 0.2 seconds across all configurations, underscoring uniform efficiency. Notably, inference speeds exhibited significant variation, with the YOLO11n-seg configuration achieving the fastest inference at 4.4 seconds per image, while the YOLO11x-seg was the slowest at 45.3 seconds, highlighting a clear trade-off between model complexity and processing speed. Post-processing times improved as model size increased, scaling from 1.8 seconds in the smallest YOLO11n-seg model to 0.7 seconds in the largest YOLO11x-seg model. Figure \ref{fig:AllREST}b and \ref{fig:AllREST}c further detail the computational complexity and training efficiency of these configurations. The number of convolution layers and GFLOPs correlated directly with model size and complexity, with the YOLO11x-seg showing the highest GFLOPs and convolution layers, indicative of its substantial computational requirements. Conversely, training times varied considerably, with YOLO11n-seg requiring only about 1.6 hours to complete 300 epochs, significantly less than the 9.9 hours required by the larger YOLO11x-seg model. This variation underscores the direct relationship between model size, parameter count, and the resources needed for training, where the smallest YOLO11n-seg model at 6.1 MB and the largest YOLO11x-seg at 124.8 MB demonstrate the spectrum of computational demand across the configurations.

Another set of parameters (e.g., convolutional layers, GFLOPs) used to assess the varying complexity of YOLO11 configurations for instance segmentation are as shown by Figure  \ref{fig:AllREST} b. The YOLO11x-seg configuration demonstrated a substantial increase in complexity and processing capability, featuring 491 convolutional layers and the highest GFLOPs at 318.5. However the computational time and cost to this configuration is much higher as this model also possessed the largest number of parameters (62 Million) as shown in Figure \ref{fig:AllREST} c, which correlates with its higher computational demands. Notably, the YOLO11x-seg model's expansive architecture contributed to its superior performance in handling more complex image segmentation tasks, likely due to its greater depth and capacity to process detailed features. This configuration, although resource-intensive, exemplifies the trade-offs between computational efficiency and model accuracy.

\subsection{Performance of YOLO11 Instance Segmentation Model against Manual Annotation}

In further assessing the model performance on the LLM-generated dataset, the YOLO11 configurations were tested against 40 such images annotated manually. Table \ref{tab:metricsLLM40imgs} shows the detailed performance measures (both box and mask metics) on the LLM-generated images against the manual annotation. The table shows that the YOLO11x-seg configuration achieved the highest performance among the tested configurations with a precision of 0.92, recall of 0.851, and mask mAP@50 of 0.92.

The results show robust instance segmentation capabilities on synthetically generated datasets as the YOLO11l-seg model, in particular, exhibited superior performance with a mask precision of 0.93 and mAP@50 of 0.92. Further evaluation against 40 manually annotated LLM-generated images, as shown in Table \ref{tab:metricsLLM40imgs}, allows for a direct assessment of model performance under conditions that simulate manual annotation processes. The metrics from this manual annotation comparison closely align with those derived from the automated annotations. The YOLO11x-seg model particularly stands out, consistently achieving high metrics in both automated and manual annotation settings, with a mask precision and mAP@50 both at 0.92. This consistency across different annotation methods underlines the models' adaptability and accuracy in real-world agricultural settings, suggesting that these configurations are not only theoretically sound but also practically effective. Such findings underscore the potential of using automatically generated and annotated datasets to train instance segmentation models, reducing reliance on labor-intensive manual annotation while maintaining high accuracy.

\begin{table*}[h]
\centering
\caption{\textbf{Evaluation of YOLO11 models on segmenting apples on LLM-generated images compared against the manual annotation of apples on those images. As discussed above, the models were trained using a dataset automatically annotated by an integrated approach using zero-shot YOLO11 and SAMv2.}}
\begin{tabular}{@{}lcccccccc@{}}
\toprule
\textbf{Model} & \multicolumn{4}{c}{\textbf{Box Metrics}} & \multicolumn{4}{c}{\textbf{Mask Metrics}} \\
& \textbf{P} & \textbf{R} & \textbf{mAP50} & \textbf{mAP50:95} & \textbf{P} & \textbf{R} & \textbf{mAP50} & \textbf{mAP50:95} \\
\midrule
YOLO11n-seg & 0.928 & 0.826 & 0.916 & 0.854 & 0.914 & 0.848 & 0.922 & 0.841 \\
YOLO11s-seg & 0.914 & 0.825 & 0.893 & 0.845 & 0.939 & 0.819 & 0.899 & 0.825 \\
YOLO11m-seg & 0.916 & 0.851 & 0.923 & 0.855 & 0.916 & 0.851 & 0.92 & 0.839 \\
YOLO11l-seg & 0.907 & 0.837 & 0.915 & 0.845 & 0.904 & 0.838 & 0.911 & 0.83 \\
YOLO11x-seg & 0.93 & 0.837 & 0.916 & 0.849 & 0.92 & 0.851 & 0.92 & 0.835 \\
\bottomrule
\end{tabular}
\label{tab:metricsLLM40imgs}
\end{table*}
\subsection{Field Validation of YOLO11 Instance Segmentation Model using Azure Camera Images}
\begin{figure*}[ht]
\centering
\includegraphics[width=0.89\linewidth]{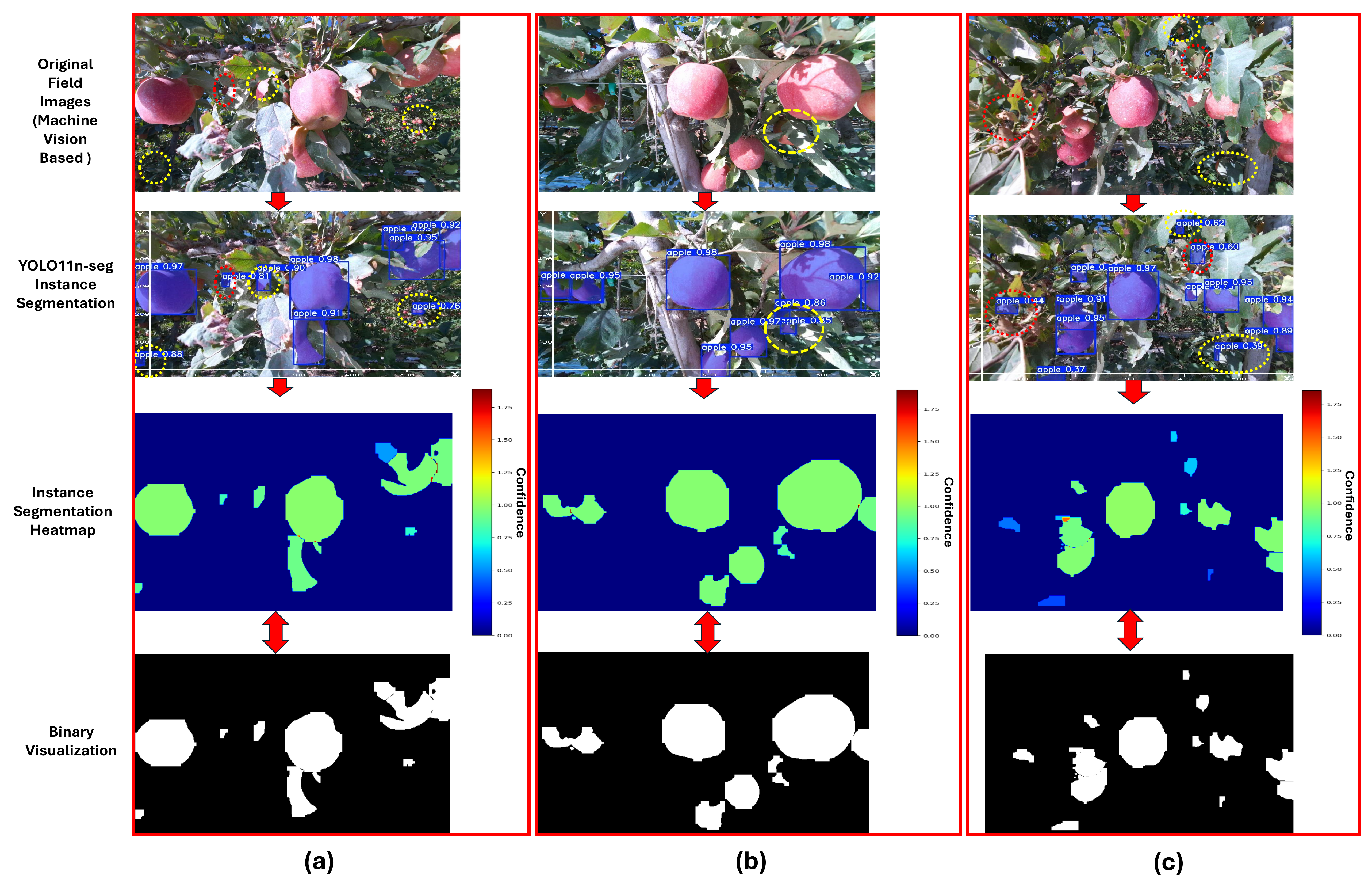}
\caption{Example results from YOLO11n-seg: (a) Shows robust detection of distant apples while also depicting false detection of shadowy foliage as apples; (b) Demonstrates the model's effectiveness in commercial orchard settings, accurately segmenting partially visible apples; and (c) Illustrates the model's capacity to segment highly occluded apples while also emphasizing its limitation of falsely detecting other background parts as apples.}
\label{fig:resultsYOLO11Machinevisiongenerated}
\end{figure*}

To assess the model's capability to be applied to actual orchard environments, a field evaluation was performed in a commercial apple orchard using an Azure machine vision camera as shown in Figure \ref{fig:resultsYOLO11Machinevisiongenerated}. A dataset of 42 test images was collected using the camera installed on a robotic platform as discussed in methods. These images were manually annotated to delineate all the apples, and were compared against the apple masks segmented by the models trained exclusively on LLM-generated, automatically annotated dataset. 
The detailed metrics calculation based on the field validation experiment are presented in Table \ref{tab:metricsFieldImages}. Among the five YOLO11 configurations tested, YOLO11m-seg demonstrated superior performance, particularly in mask metrics, where it achieved a mask precision of 0.902 and a mask mAP@50 of 0.833 (Table \ref{tab:metricsFieldImages}). This model configuration (as well as other configurations) showed good adaptability to real orchard environments, effectively delineating apple instances in images collected by a machine vision camera despite the inherent challenges posed by natural orchard environments. The reasonable high precision and recall demonsrated the potential of YOLO11m-seg for practical agricultural applications, confirming its efficacy in handling complex visual data enabling automated agricultural monitoring and interventions. 

\begin{table*}[h]
\centering
\caption{\textbf{Evaluation of instance segmentation metrics for various configurations of YOLO11 model. The models were trained using a dataset automatically annotated by SAM, illustrating how well each configuration performs in recognizing and delineating apples in field-collected  images collected by Microsoft Azure Kinect Camera. This underscores the robust adaptability of these models in real-world agricultural environments.}}
\begin{tabular}{@{}lcccccccc@{}}
\toprule
\textbf{Model} & \multicolumn{4}{c}{\textbf{Box Metrics}} & \multicolumn{4}{c}{\textbf{Mask Metrics}} \\
& \textbf{P} & \textbf{R} & \textbf{mAP50} & \textbf{mAP50:95} & \textbf{P} & \textbf{R} & \textbf{mAP50} & \textbf{mAP50:95} \\
\midrule
YOLO11n-seg & 0.889 & 0.714 & 0.848 & 0.760 & 0.889 & 0.714 & 0.848 & 0.703 \\
YOLO11s-seg & 0.827 & 0.717 & 0.823 & 0.747 & 0.849 & 0.700 & 0.820 & 0.689 \\
YOLO11m-seg & 0.886 & 0.696 & 0.828 & 0.742 & 0.902 & 0.700 & 0.833 & 0.681 \\
YOLO11l-seg & 0.871 & 0.706 & 0.826 & 0.746 & 0.878 & 0.706 & 0.822 & 0.687 \\
YOLO11x-seg & 0.817 & 0.720 & 0.833 & 0.757 & 0.817 & 0.720 & 0.830 & 0.705 \\
\bottomrule
\end{tabular}
\label{tab:metricsFieldImages}
\end{table*}

Figure \ref{fig:resultsYOLO11Machinevisiongenerated} presents the sample results of the YOLO11 instance segmentation models on orchard images acquired using a Microsoft Azure camera. Figure \ref{fig:resultsYOLO11Machinevisiongenerated} is composed of four layers to provide a comprehensive view of the model's application in real-world settings. The topmost layer captures the original orchard imagery, succeeded by the YOLO11n model’s detection and segmentation outcomes, visually demonstrated using automatically generated images and annotations. Subsequent layers include a heatmap depicting segmentation intensity and a binary image that delineates the segmented apples, illustrating the model's effectiveness in complex environments. This structured display highlights the model's proficiency in identifying distant apples despite their size or relative distance. Yet, challenges persist, as indicated by the misidentification of shadow-cast foliage as apples in red-circled areas, suggesting areas for future refinement to improve accuracy under varied orchard conditions. 

Figure \ref{fig:resultsYOLO11Machinevisiongenerated}b vividly demonstrates the YOLO11 instance segmentation model's effectiveness within a commercial orchard setting. In the yellow dotted regions, the model excelled by accurately detecting and segmenting apples that were only 5-8\% (approx) visible, showcasing its exceptional ability to recognize and delineate apples even when minimally exposed. This performance highlights the success of the model in navigating and processing complex agricultural scenes, reinforcing the robustness of the training on LLM-generated and SAM-annotated datasets. In contrast, Figure \ref{fig:resultsYOLO11Machinevisiongenerated}c presents a mixed outcome where the model's strengths and limitations are further explored. The yellow circled area illustrates the model's proficiency in identifying and segmenting an apple in highly occluded scenarios where only  3-5\% (approx) of the fruit is visible. This example underscores the model's advanced capability to handle intricate and obstructed scenes effectively. However, the same figure also points to areas needing improvement, as indicated by the red-circled regions. Here, the model erroneously detected and segmented dry leaves and canopy foliage, mistaking them for apples due to their brownish color. These misclassifications highlight the challenges faced by the model in differentiating between apples and similarly colored non-target elements, suggesting a need for further refinement to enhance its accuracy in complex visual environments.

\subsubsection{Evaluation of Image Processing Speeds}

\begin{figure*}[ht]
\centering
\includegraphics[width=0.89\linewidth]{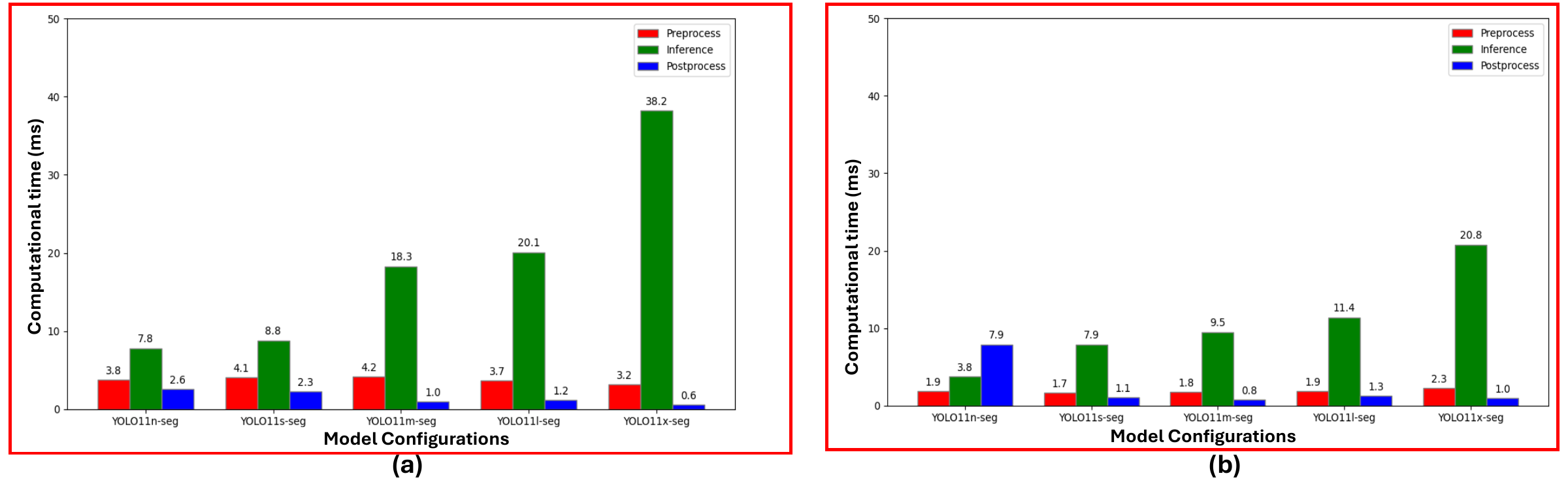}
\caption{(a) Computational time taken by various YOLO11 model configurations on processing LLM-generated images for the 40 validation images ; and (b) the same for 42 orchard images acquired with a machine vision sensor (Microsoft Azure Camera) }
\label{fig:InferenceValidation}
\end{figure*}
The computational speed of all configurations of YOLO11 instance segmentation model with 40 LLM-generated images and 42 camera-acquired images are detailed in Figure \ref{fig:InferenceValidation}. The figure shows that computational speed across various model configurations varied substantially between LLM-generated and real camera-generate images. Among the configurations tested, YOLO11x-seg consistently outperformed others in terms of inference speed, though it exhibited the highest speeds due to its more extensive computational requirements. Specifically, for LLM-generated images, preprocessing time ranged from 3.2 ms to 4.2 ms where YOLO11x-seg was the fastest at 3.2 ms. Inference times for this configuration was the highest at 38.2 ms, indicating more intensive computational need for this model, while post-processing computation was swift at 0.6 ms. Similarly, for images collected using the Microsoft Azure Kinect camera, YOLO11x-seg demonstrated optimal performance in pre-processing, and inference, albeit at a slightly lower speeds compared to the same for the LLM-generated dataset. Pre-processing times varied slightly, with YOLO11x-seg showing a slightly increased time of 2.3 ms. Inference times were notably faster across all configurations, with YOLO11x-seg again taking the longest at 20.8 ms, which was still substantially faster than its speed on synthetic images. Post-processing speed for this configuration was 1.0 ms. 

\newpage 
\section{Discussion}
\begin{figure*}[ht!]
\centering
\includegraphics[width=0.85\linewidth]{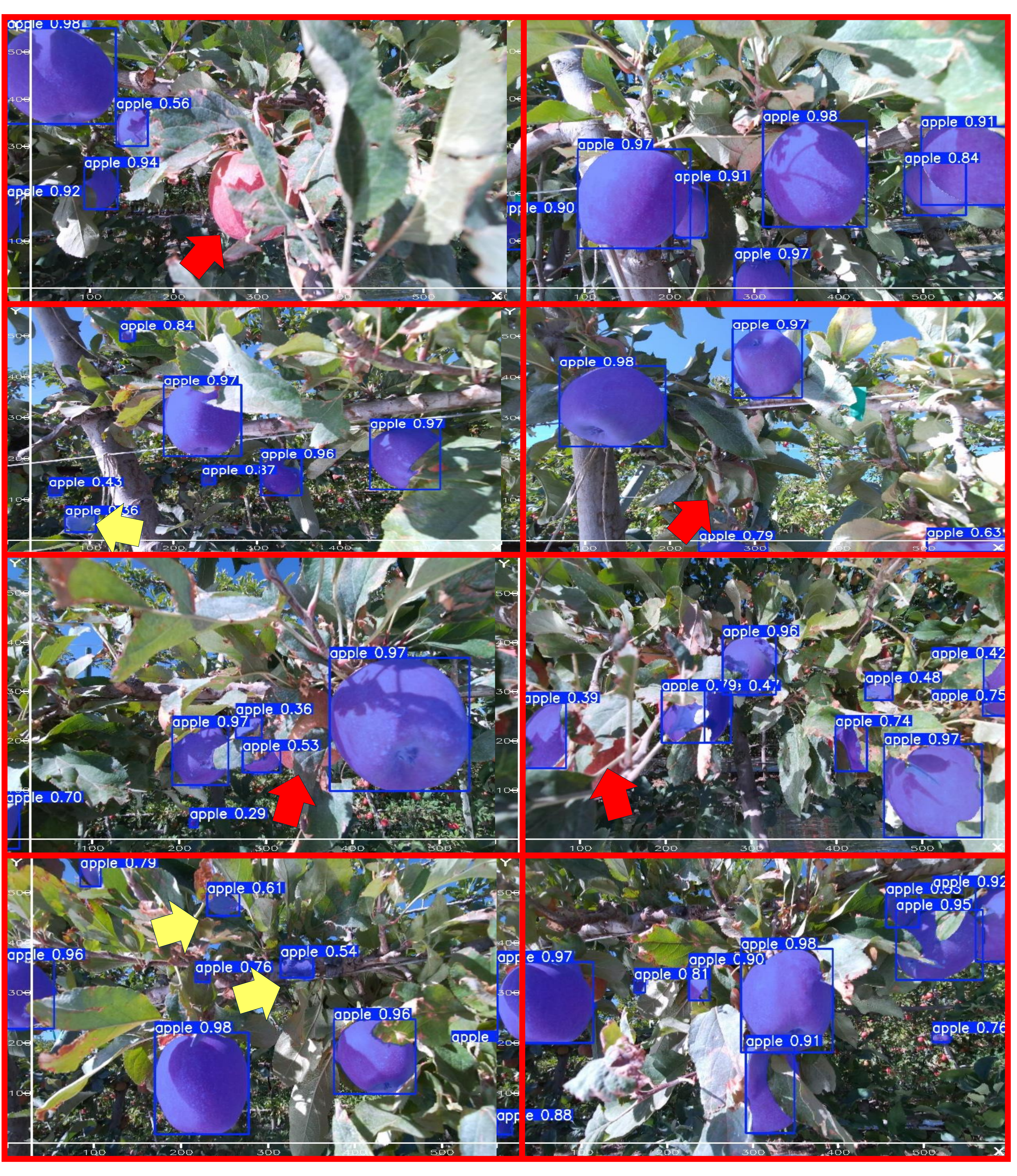}
\caption{Showing the examples of the efficacy of our zero-shot learning-based instance segmentation model on commercial apple orchards, with a Microsoft Azure machine vision camera. This figure demonstrates the model's performance, trained solely on LLM-generated and SAM-annotated synthetic datasets. Red arrows highlight regions where occluded apples were not segmented due to complex environmental factors, while yellow arrows show false positive segmentations of canopy foliage as apples. These visual outcomes emphasize the model’s potential and the need for enhanced data realism and volume to improve accuracy in real-world applications.}
\label{fig:ExamplesFieldImages}
\end{figure*}

In \ref{fig:ExamplesFieldImages}, the outcomes of applying our zero-shot learning-based instance segmentation model to commercial apple orchards are illustrated, with images captured via a robotic platform equipped with a Microsoft Azure machine vision camera. This visual representation showcases different examples of field-level instance segmentation, performed by a model trained exclusively on synthetic datasets generated by LLM and automatically annotated through SAM-YOLO11 fusion. This approach marks a significant departure from traditional methods that rely heavily on sensor-based field data collection and manual annotations.

The images in the figure, annotated with red and yellow arrows, highlight the model's capabilities and limitations. Red arrows indicate areas where the model failed to segment occluded apples due to complex environmental conditions, suggesting a need for richer datasets. This study, limited to 501 synthetic images, highlights the potential for expanding data generation using LLM capabilities, potentially integrating multimodal inputs such as textual prompts and voice. The yellow arrows denote false segmentations where canopy foliage was incorrectly identified as apples, underscoring the importance of generating more realistic training images to improve model accuracy.

Our approach contrasts sharply with recent advancements in few-shot and zero-shot learning that still require some form of initial data handling or manual intervention. For instance, studies like those by \cite{keaton2023celltranspose} and \cite{wang2022dynamic} have made significant strides in using few-shot learning to reduce data dependence, but they still rely on limited data collection and manual annotations. Similarly, zero-shot models like those proposed by \cite{zheng2021zero} and \cite{shin2023zero} eliminate the need for labeled data for unseen classes but have not been applied to completely synthetic datasets for training, as in our study.

This research extends the boundary of what is possible with zero-shot learning in agriculture by demonstrating that robust instance segmentation models can be developed and deployed without any field data collection or manual labeling, setting a new standard for scalability and adaptability in agricultural AI applications. The integration of LLM-generated datasets and zero-shot learning significantly reduces the time, cost, and logistical challenges typically associated with model training and deployment, offering a novel and effective solution for real-world agricultural monitoring and intervention tasks.

Our study's uniqueness and strength lie in its complete independence from physical data collection and manual annotation processes, distinguishing it from the incremental improvements reported in existing literature. While the works of \cite{nguyen2022ifs}, \cite{zheng2021zero}, and others represent critical steps towards reducing labor in model training, they have not achieved the level of autonomy in model preparation and deployment that this research has demonstrated. By leveraging synthetic data and automatic annotations, our model not only simplifies the preparatory phases of machine learning but also enhances the feasibility of deploying advanced AI solutions across varied and resource-constrained agricultural settings, embodying the next level of innovation in agricultural technology.

Additionally Our approach introduces a novel paradigm that transcends traditional data augmentation techniques commonly employed in training deep learning models. Unlike simple geometric or photometric transformations \cite{taylor2018improving, shorten2019survey, mumuni2022data}, synthetic data generation through LLMs offers a more expansive and diverse set of training scenarios, potentially covering a broader range of object appearances and conditions not typically available in existing datasets \cite{sapkota2024synthetic, sapkota2024multi}. This method not only enhances the robustness of the model against varied real-world conditions but also reduces the bias inherent in limited dataset compositions.

Furthermore, the capability to quickly adapt models to new agricultural challenges without the prerequisite of localized data collection presents strategic advantages \cite{fuentes2024transformative, huo2024mapping}, especially in emerging markets and in scenarios demanding rapid responses to agricultural threats \cite{tripathi2023farmer}. This flexibility could dramatically shorten response times in managing outbreaks of pests or crop diseases, which is crucial for preventing widespread damage in vulnerable regions \cite{shafiq2024climate, subedi2023impact}. Furthermore, the application of zero-shot learning models in these contexts underscores the potential for AI to deliver high-impact solutions with minimal logistical overhead, thereby democratizing access to cutting-edge technology across varying economic landscapes.

\section{Conclusion and Future}

This study successfully developed a robust deep learning model for instance segmentation in commercial orchard environments without the traditional reliance on sensor data collection or manual annotations. Through the integration of a Large Language Model (LLM) with the Segment Anything Model (SAM) and a zero-shot YOLO11 base model, we generated and annotated synthetic images of apple orchards. This innovative approach facilitated the training of the YOLO11 model, which demonstrated high accuracy when validated against a manually annotated commercial orchard dataset. The results underscored the potential of synthetic datasets and zero-shot learning to significantly enhance the scalability and efficiency of AI deployments in agriculture. Here are the major highlights from our findings: 
\begin{itemize}
    \item Zero-Shot Base Model Performance: The zero-shot YOLO11 base model adeptly detected apples within the LLM-generated images, demonstrating a mask precision of 0.92, recall of 0.851, and a mask mAP@50 of 0.92. These metrics affirm the model's high accuracy in segmentation tasks. 
    \item Automatic Annotation Accuracy: The automatically generated annotations showcased impressive performance metrics, with a Dice Coefficient of 0.9513 and an IoU of 0.9303, highlighting the precise overlap and accuracy of the mask annotations. 
    \item Training with LLM-Generated Datasets: All YOLO11 configurations were trained solely on these synthetic datasets, proving effective in adapting the models to recognize and delineate apples accurately within digitally created orchard environments. 
    \item Validation on LLM-Generated Images: The models were validated on 40 LLM-generated images, and the YOLO11l-seg model outshone others by achieving the highest mask precision and overall mAP@50 metrics, indicating superior performance in delineating apple instances accurately. 
    \item Field-Level Validation with Real Images: In a further test, 42 field-collected images were used for validation. The YOLO11m-seg configuration demonstrated exceptional performance on real-world data, particularly in mask metrics, achieving a mask precision of 0.902 and a mask mAP@50 of 0.833.
\end{itemize}
In the future, this approach could be adapted to develop custom models for a wide array of objects, not limited to different types of fruits like apples, but also for any objects across various industries where instance segmentation is crucial. By leveraging synthetic image generation and zero-shot learning, we can significantly streamline the development of deep learning models, making the technology accessible and adaptable for diverse and resource-constrained environments.

\textbf{Authors' Contribution:} 
Ranjan Sapkota : conceptualization, data curation, software, methodology, validation,  writing original draft. Achyut Paudel: software, review and editing.  Manoj Karkee (correspondence)  review, editing and overall funding to supervisory. 

Our other Research on \cite{sapkota2024immature}, \cite{sapkota2024synthetic}, \cite{sapkota2024integrating}, \cite{sapkota2024yolo11}, \cite{sapkota2024yolov10}, \cite{meng2025yolov10}, \cite{sapkota2023creating}, \cite{sapkota2024multi}, \cite{churuvija2025pose}, \cite{sapkota2024comparing}, \cite{khanal2023machine}, \cite{sapkota2025image}, \cite{sapkota2025comprehensive}, \cite{sapkota2025comprehensiveanalysistransparencyaccessibility}, \cite{sapkota2025yolov12genesisdecadalcomprehensive}, \cite{sapkota2025Improved}. 
\bibliography{references.bib}
\begin{IEEEbiography}[{\includegraphics[width=1in,height=0.85in,clip,keepaspectratio]{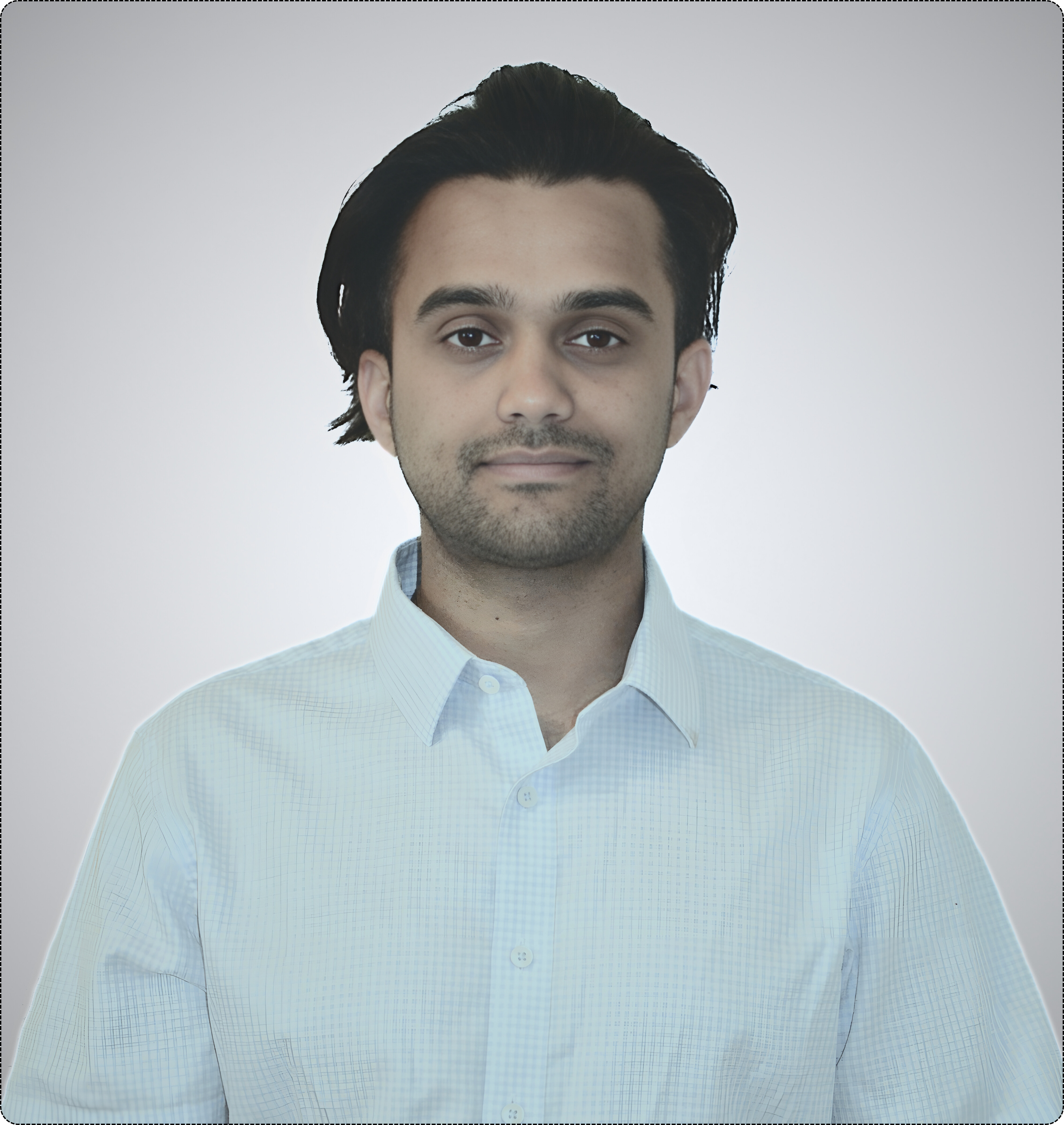}}]{Ranjan Sapkota} ( Member, IEEE) obtained his B.Tech degree in Electrical and Electronics Engineering from Uttarakhand Technical University, India in 2019. He then pursued his MS in Agricultural and Biosystems Engineering from North Dakota State University from 2020 to 2022. Currently, Mr. Sapkota is a Ph.D. student at Washington State University in the Department of Biological Systems Engineering and Center for Precision and Automated Agricultural Systems. His research primarily focuses on Automation and Robotics for Agriculture, utilizing Artificial Intelligence, Large Language Models, Machine Vision, Robot Manipulation Systems, Deep Learning, Machine Learning and Generative AI Technologies.
\end{IEEEbiography}
\begin{IEEEbiography}
[{\includegraphics[width=1in,height=0.85in,clip,keepaspectratio]{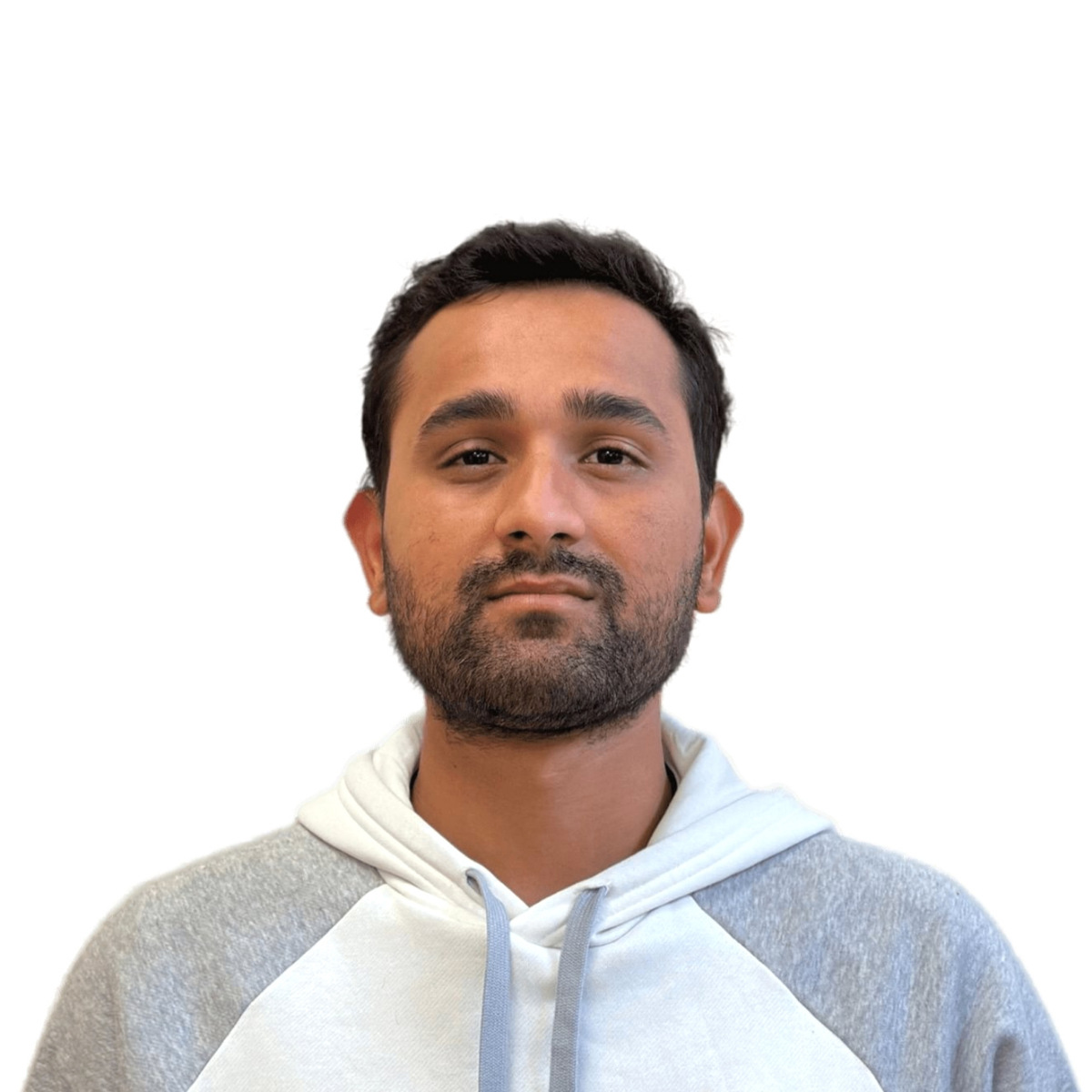}}]{Achyut Paudel} is a Ph.D. candidate in Agricultural Engineering at Washington State University, where he also earned his M.S. in Mechanical Engineering. He holds a B.S. in Mechanical Engineering from Tribhuvan University, Nepal. Currently serving as a Machine Learning/Computer Vision Research Engineer at Orchard Robotics, Paudel focuses on automating various tasks in orchards and vineyards. His research leverages machine vision and artificial intelligence to enhance agricultural practices.
\end{IEEEbiography}
\begin{IEEEbiography}
[{\includegraphics[width=1in,height=0.85in,clip]{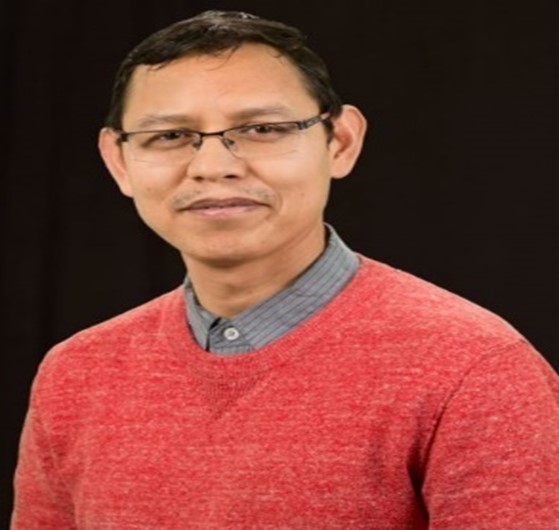}}]{Prof. Dr. Manoj Karkee} obtained his BS in Computer Engineering from Tribhuwan University in 2002. He pursued his MS in Remote Sensing and Geographic Information Systems at Asian Institute of Technology, Thailand, and earned his Doctorate in Agricultural Engineering and Human-Computer Interaction from Iowa State University in 2009. Dr. Karkee currently serves as the Professor and Director of the Center for Precision and Automated Agricultural Systems at Washington State University. His research focuses on agricultural automation and mechanization programs, with an emphasis on machine vision-based sensing, automation and robotic technologies for specialty crop production.
\end{IEEEbiography}
\bibliographystyle{ieeetr}

\end{document}